\pdfoutput=1
\documentclass[10pt,journal,compsoc]{IEEEtran}
\usepackage{graphicx}
\usepackage{caption}
\usepackage{bigstrut}
\usepackage{array}
\usepackage{booktabs}
\usepackage{tabularx}
\usepackage[table]{xcolor}
\usepackage{longtable}
\usepackage{hyperref}
\usepackage{mathtools}
\usepackage{amssymb}
\usepackage{amsthm}
\hypersetup{
    colorlinks=true,
    linkcolor=magenta,
    filecolor=magenta,      
    urlcolor=blue,
    pdftitle={Survey of Textual QA Benchmarks},
    pdfpagemode=FullScreen,
    }

\begin{document}


\title{More Than Reading Comprehension: A Survey on Datasets and Metrics of Textual Question Answering}

\author{Yang~Bai, \ 
        Daisy~Zhe~Wang
\IEEEcompsocitemizethanks{\IEEEcompsocthanksitem Y. Bai and D.Z.Wang are with the Department
of Computer and Information of Science and Engineering, University of Florida, Gainesville,
FL, 32611.\protect\\

E-mail: baiyang94@ufl.edu}
}



\IEEEtitleabstractindextext{%
\begin{abstract}

Textual Question Answering (QA) aims to provide precise answers to users' questions in natural language using unstructured data. One of the most popular approaches to this goal is machine reading comprehension(MRC). In recent years, many novel datasets and evaluation metrics based on classical MRC tasks have been proposed for broader textual QA tasks. In this paper, we survey 47 recent textual QA benchmark datasets and propose a new taxonomy from an application point of view. In addition, We summarize 8 evaluation metrics of textual QA tasks. Finally, we discuss current trends in constructing textual QA benchmarks and suggest directions for future work.

\end{abstract}

}

\maketitle

\IEEEdisplaynontitleabstractindextext
\IEEEpeerreviewmaketitle

\IEEEraisesectionheading{\section{Introduction}\label{sec:introduction}}

\begin{figure*}[t]
    \centering
    \captionsetup{justification=centering}
    \includegraphics[scale=0.8]{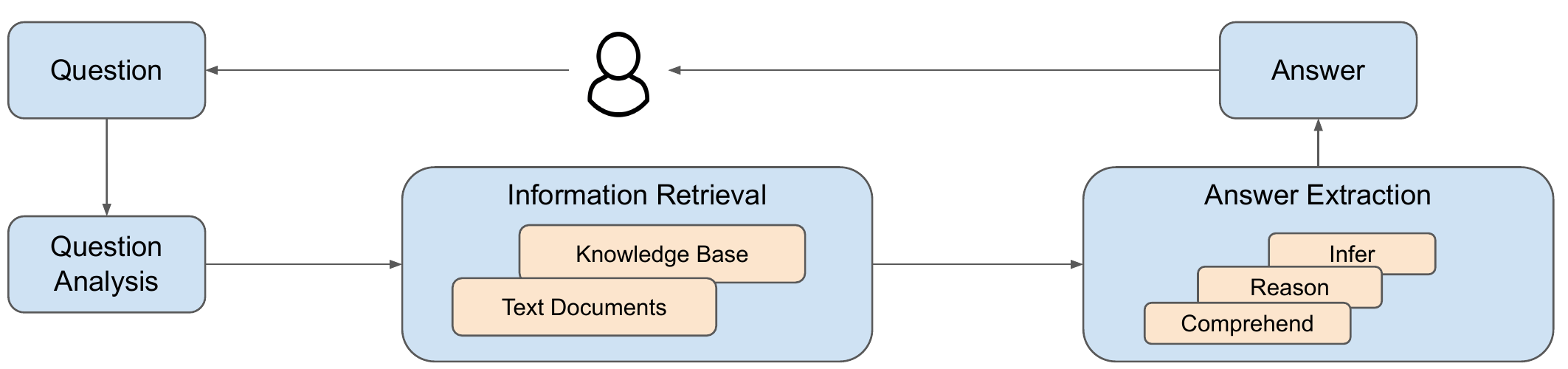}
    \caption{Generic QA system architecture.} 
    \label{fig:qa_general_architecture}
\end{figure*}

\IEEEPARstart{Q}{uestion} Answering(QA) is an area of research in the field of natural language processing that dates back to the 1960s\cite{Green1961BaseballAA}. In contrast to traditional information retrieval(IR) which aims to retrieve relevant documents given a query, QA aims to give precise answers to users’ questions in natural language\cite{Voorhees2000BuildingAQ}. This is also a higher goal that has been motivating current search engines. Based on the structure of data, QA can be divided into Knowledge Base (KB) QA and textual QA. KBQA leverages a predefined structured KB (e.g., Freebase\cite{Bollacker2008FreebaseAC}, YAGO\cite{Suchanek2007YagoAC}) to answer questions, while textual QA mines answers from unstructured text data (e.g., Wikipedia documents, news articles, etc.). The architecture of a generic QA system is shown in Figure~\ref{fig:qa_general_architecture}.

Machine Reading Comprehension (MRC), a popular textual QA application has been well studied. Early MRC research focused on answering relatively simple questions requiring no complex reasoning, given a short context passage containing an explicit answer. Large-scale classical MRC benchmarks aided the development of deep learning-based language models (e.g., BERT\cite{devlin-etal-2019-bert}, GPT-3\cite{Brown2020LanguageMA}, BART\cite{lewis-etal-2020-bart}, T5\cite{Raffel2020ExploringTL}, etc.) and QA systems based on them. Already, deep MRC systems have matched or beaten human performance on early classical MRC benchmarks\cite{Dzendzik2021EnglishMR}. Based on these early efforts, recently, many new datasets for various application scenarios of textual QA were proposed. They are either a form of MRC, e.g., conversational QA\cite{Saeidi2018InterpretationON}\cite{Choi2018QuACQA}\cite{Reddy2019CoQAAC}, multi-hop QA\cite{Yang2018HotpotQAAD}\cite{Ho2020ConstructingAM}\cite{Chen2020HybridQAAD}, long-form QA\cite{Fan2019ELI5LF}\cite{zhu2020question}\cite{Soleimani2021NLQuADAN}, and cross-language QA\cite{liu-etal-2019-xqa}\cite{artetxe-etal-2020-cross}\cite{lewis-etal-2020-mlqa}, which are referred to as novel MRC in this paper; or in a related form, such as open-domain QA(ODQA)\cite{chen-etal-2017-reading}\cite{Lee2019LatentRF}\cite{Wang2018R3RR}\cite{Petroni2021KILTAB}, or commonsense QA\cite{Zellers2018SWAGAL}\cite{chen-etal-2019-codah}\cite{Talmor2019CommonsenseQAAQ}\cite{Huang2019CosmosQM}\cite{Zhang2018ReCoRDBT}. These novel challenging subtasks of textual QA have individually received a great deal of research attention in recent years and spurred improvements in existing language understanding and textual QA systems. However, sometimes this wide array of datasets for various application scenarios can be confusing to uninitiated researchers. It motivates us to provide a comprehensive and comparative survey of existing public benchmarks to help researchers new to the field quickly understand the current trends and challenges of textual QA.

\begin{figure*}[t]
    \centering
    \captionsetup{justification=centering}
    \includegraphics[scale=0.35]{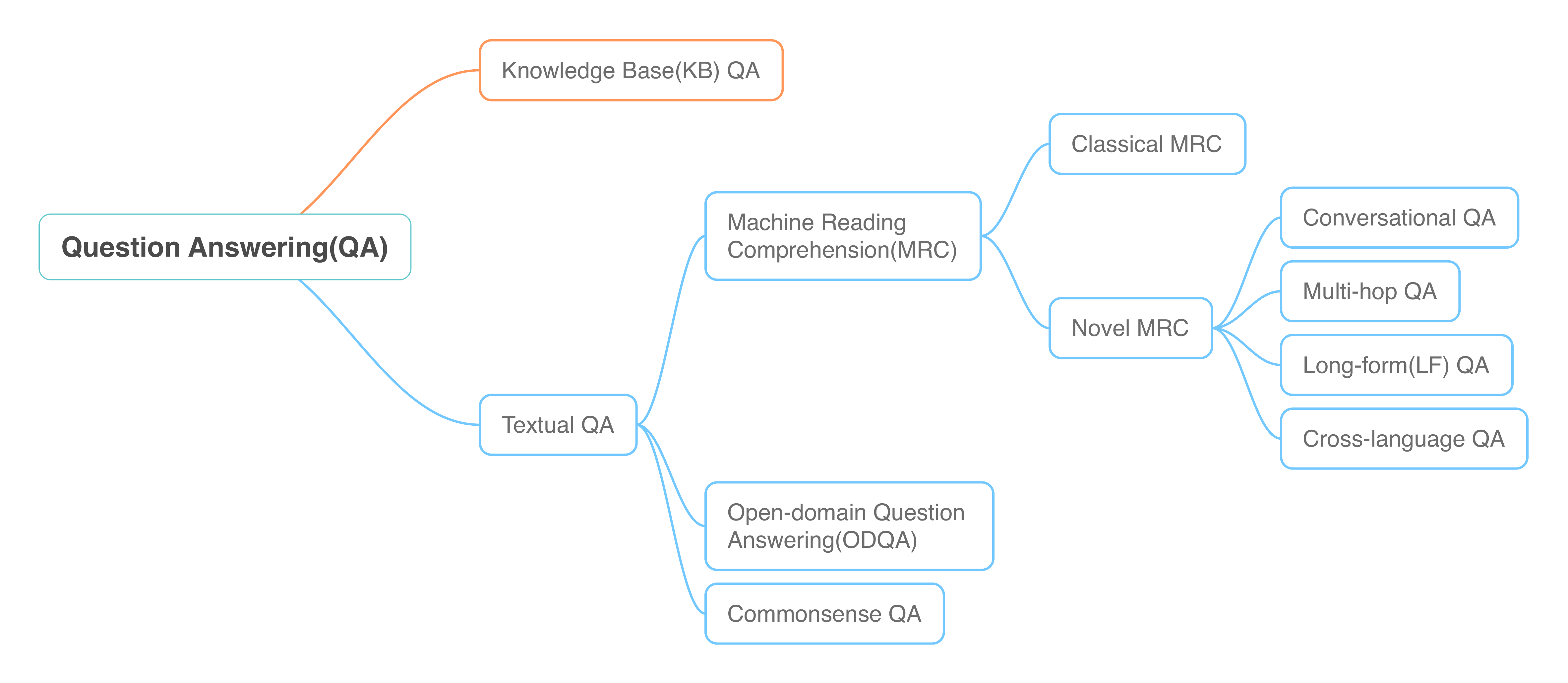}
    \caption{Taxonomy of textual QA by application scenarios.} 
    \label{fig:qa_taxonomy}
\end{figure*}

Our contributions are summarized as follows:
\begin{enumerate}
    \item A new taxonomy for textual QA from a novel application scenarios point of view is proposed as shown in Figure~\ref{fig:qa_taxonomy}.
    \item It covers 7 major subtasks of textual QA: classical MRC, 4 novel MRC tasks (including conversational QA, multi-hop QA, long-form QA, and cross-language QA), ODQA, and commonsense QA.
    \item Comparisons are applied over multiple dataset facets.
    \item Detailed statistics of each dataset are provided.
    \item Evaluation metrics of each benchmark dataset are described.
    \item Finally, we discuss current trends and future directions in textual QA to help people focus on unsolved challenging tasks and propose new more challenging datasets.
\end{enumerate}

\subsection{Related Surveys}
Similar works have been done over textual QA benchmarks previously \cite{Cambazoglu2020ARO}\cite{Zeng2020ASO}\cite{Dzendzik2021EnglishMR}\cite{Zhu2021RetrievingAR}\cite{Zaib2021ConversationalQA}. But each of them only covers a few subtasks of textual QA. For example, \cite{Cambazoglu2020ARO} carefully reviewed the construction methods and the leaderboard information of 22 public MRC datasets; \cite{Zeng2020ASO} provides detailed statistics and analysis over 54 MRC datasets; \cite{Zhu2021RetrievingAR} mainly focused on ODQA tasks; \cite{Dzendzik2021EnglishMR} is a comprehensive survey on English MRC datasets and metrics; \cite{Zaib2021ConversationalQA} is a detailed survey over the Conversational QA tasks. The major difference between our work and previous works is its breadth and granularity. This survey covers three major types of textual QA tasks: MRC, ODQA, and commonsense QA. And we further differentiate the MRC datasets as classical MRC tasks and four novel MRC tasks: conversational QA, multi-hop QA, long-form QA, and cross-language QA.

\subsection{Structure of the survey:}
The rest of the paper is structured as follows: 

In section~\ref{sec:MRC}, we reviewed 31 MRC benchmark datasets in total. We first introduced the traditional categorization of MRC datasets. Then following this traditional categorization, we reviewed 18 classical MRC datasets. Next, we reviewed 13 novel MRC datasets of 4 different types: conversational QA, multi-hop QA, long-form QA, and cross-language QA. The review of each dataset covers its construction method and statistical information. The information and statistics are summarized in tables.

In section~\ref{sec:ODQA}, we summarized and reviewed 11 popular textual QA datasets that were used in an Open-Domain Question Answering(ODQA) setting in recent papers. We first summarized the datasets and metrics that were selected to train or test the ODQA systems proposed in recent papers. Then we reviewed datasets, such as Facebook AI's KILT, that were integrated into a single ODQA benchmark. Multiple feature facets are summarized and compared.

In section~\ref{sec:Commonsense QA}, we present a spatial type of textual QA task, Commonsense Question Answering. We discussed the uniqueness of this task compared to MRC and ODQA. In total, we reviewed 5 benchmark datasets of this kind. Straightforward examples of each dataset are given along with figures borrowed from the original papers. Information on each dataset is summarized in a table.

In section~\ref{sec:Metrics}, we summarized the metrics that were used in each textual QA task covered in this paper. A description of and the equations for each metric are provided. The distribution of metrics by answer type is shown in a chart. 

In section~\ref{sec:Trends&Directions}, we summarized the trends in constructing textural QA tasks and propose future directions of textual QA benchmark research.

Finally, in section~\ref{sec:Conclusion}, we conclude our survey.

\section{Machine Reading Comprehension}\label{sec:MRC}

The goal of machine reading comprehension(MRC) is to develop systems capable of understanding natural language. The system's comprehension is measured by testing its ability to answer questions based on secondary text. Motivated by the development of deep-learning language models(BERT\cite{devlin-etal-2019-bert}, BART\cite{lewis-etal-2020-bart}, GPT-3\cite{Brown2020LanguageMA}, T5\cite{Raffel2020ExploringTL}, XLNet\cite{Yang2019XLNetGA}, etc) where large-scale benchmark datasets played a crucial role, many new MRC datasets have been proposed.  

\begin{figure}
    \centering
    \includegraphics[width=\linewidth]{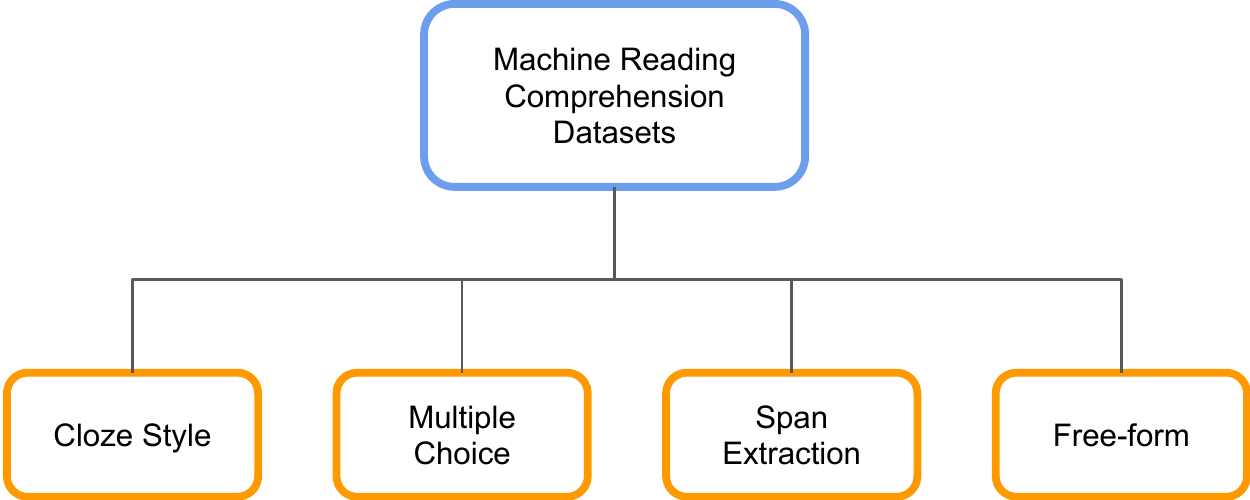}
    \caption{Traditional categorization of MRC datasets.}
    \label{fig:MRC_cate}
\end{figure}

MRC datasets are commonly classified by annotation style into four categories: cloze style, multi-choice style, span extraction style, and free-form style\cite{Zeng2020ASO}\cite{Zaib2021ConversationalQA}. See Figure~\ref{fig:MRC_cate}.

\begin{itemize}
  \item \textbf{Cloze Style} datasets are usually constructed programmatically by taking a part of a sentence in a passage as the answer, the rest of the sentence as the question, and the context consists of the sentences close to the question sentence. Given context content, the MRC systems are required to complete the question sentence with a word or phrase from a predefined dictionary\cite{hermann2015teaching}\cite{cui-etal-2016-consensus} or from a set of selections\cite{Hill2016TheGP}\cite{Bajgar2016EmbracingDA}.\\
  
  \item \textbf{Multi-choice Style} datasets are constructed by hiding the correct answer into a set of distracting answers to a question based on the context. Systems are required to pick the correct answer from the misleading candidates given the question and the context.\\
  
  \item \textbf{Span Extraction Style} datasets are annotated by using a span of words in the context passage as the answer. Systems need to predict the span of words that match the annotation given the question and the context.\\
  
  \item \textbf{Free-form Style} datasets use manually created answers in their annotation which may not exactly match the words in the context. Systems are allowed to generate any form of text as the answer, which usually requires more complex metrics in evaluation.
\end{itemize}

In this paper, we further categorized MRC datasets as classical MRC or novel MRC from an application scenario perspective using the following novel features as criteria:

\begin{enumerate}
  \item If conversational turns are involved (conversational QA). 
  \item If multi-step reasoning is required (multi-hop QA).
  \item If the answers are longer than one sentence (long-form QA).
  \item If it supports cross-language question answering (cross-language QA).
\end{enumerate}

Table~\ref{tab:MRC} compares all MRC datasets. The datasets of different annotation forms are shown in different colors. Table~\ref{tab:MRC_stats} gives statistics of each MRC datasets.

\begin{table*}[htbp]
    \centering
    \captionsetup{justification=centering}
    \caption{Comparison of Machine Reading Comprehension datasets.} \label{tab:MRC}%
    \setlength{\tabcolsep}{1pt} 
    \renewcommand{\arraystretch}{1.5} 
    \begin{tabular}{@{\extracolsep{1pt}}*1l@{} @{\extracolsep{1pt}}*7c@{} @{\extracolsep{1pt}}*1r@{}}
        \toprule
        &  &  &  & \multicolumn{3}{c}{\textbf{Source}}\\\cline{5-7}
        \textbf{Dataset} & \textbf{Year} & \textbf{Language} & \textbf{Context} & \textbf{Passages} & \textbf{Questions} & \textbf{Answers} & \textbf{Answer Type} & \textbf{Metrics}\\
        \hline
        &&&&Classical MRC\\
        \midrule
         CNN/DM\cite{hermann2015teaching} & 2015 & EN & News & CNN, Daily Mail & \cellcolor{yellow!25}Synthetic Cloze & \cellcolor{yellow!25}Synthetic Cloze & Fill in entity & Accuracy\\
        
        CBT\cite{Hill2016TheGP} & 2015 & EN & Children's books & Project Gutenberg & \cellcolor{yellow!25}Synthetic Cloze & \cellcolor{yellow!25}Synthetic Cloze & Fill in word & Accuracy\\
        
        BookTest\cite{Bajgar2016EmbracingDA} & 2016 & EN & Books & Project Gutenberg & \cellcolor{yellow!25}Synthetic Cloze & \cellcolor{yellow!25}Synthetic Cloze & Fill in word & Accuracy\\
        
        HLF-RC\cite{cui-etal-2016-consensus} & 2016 & ZH & Story, News & People Daily/ & \cellcolor{yellow!25}Synthetic Cloze & \cellcolor{yellow!25}Synthetic Cloze & Fill in word & Accuracy\\&&&&Childern's Fairy Tale\\
        
        \midrule
        
        MCTest\cite{richardson-etal-2013-mctest} & 2013 & EN & Fictional Stories & Crowdsourced & Crowdsourced & Crowdsourced & \cellcolor{purple!25}Multi. Choices & Accuracy\\
        
        
        MovieQA\cite{Tapaswi2016MovieQAUS} & 2016 & EN & Movie & Movie stories & Crowdsourced & Crowdsourced & \cellcolor{purple!25}Multi.Choice & Accuracy\\
        
        RACE\cite{Lai2017RACELR} & 2017 & EN & English Exams & English Exams & English Exams & English Exams & \cellcolor{purple!25}Multi.Choice & Accuracy\\
        
        \midrule
        
        SQuAD1.1\cite{Rajpurkar2016SQuAD1Q} & 2016 & EN & Wikipedia & Wikipedia passages & Crowdsourced & Crowdsourced & \cellcolor{blue!25}Span of words & F1, EM\\
        
        SQuAD2.0\cite{Rajpurkar2018KnowWY} & 2018 & EN & Wikipedia & Wikipedia passages & Crowdsourced & Crowdsourced & \cellcolor{blue!25}Span of words & F1, EM\\
        
        NewsQA\cite{Trischler2017NewsQAAM} & 2017 & EN & News & CNN articles & Crowdsourced & Crowdsourced & \cellcolor{blue!25}Span of words & F1, EM\\
        
        SearchQA\cite{Dunn2017SearchQAAN} & 2017 & EN & Jeopardy! & Web doc. & J!Archive & J!Archive & \cellcolor{blue!25}Span of words & Accuracy, F1\\
        
        TriviaQA-Web\cite{Joshi2017TriviaQAAL} & 2017 & EN & Trivia & Web doc & Trivia Websites & Trivia Websites & \cellcolor{blue!25}Span of words & F1, EM\\
        
        TriviaQA-Wiki\cite{Joshi2017TriviaQAAL} & 2017 & EN & Trivia & Wikipedia & Trivia Websites & Trivia Websites & \cellcolor{blue!25}Span of words & F1, EM\\
        
        NaturalQS\cite{Kwiatkowski2019NaturalQA} & 2019 & EN & Wikipedia & Wikipedia & User search logs & Crowdsourced & \cellcolor{blue!25}Span of words & F1\\

        \midrule
        
        MS-MARCO\cite{nguyen2016ms} & 2016 & EN & Web search & Web doc. & User search logs & Crowdsourced & \cellcolor{green!25}Free-form & ROUGE-L, BLEU-1\\
        
        NarrativeQA\cite{Kocisk2018TheNR} & 2017 & EN & Fictional Stories & Movie scrips\&Books & Crowdsourced & Crowdsourced & \cellcolor{green!25}Free-form & ROUGE-L, BLEU-1,\\
        &&&&&&&& BLEU-4, METEOR\\
        
        DuReader\cite{He2018DuReaderAC} & 2017 & ZH & Web search & Web doc./ & User search logs & Crowdsourced & \cellcolor{green!25}Free-form & ROUGE-L, BLEU-4\\
        &&&& Baidu Zhidao\\
        
        TweetQA\cite{Xiong2019TWEETQAAS} & 2019 & EN & Tweets & CNN/NBC websites & Crowdsourced & Crowdsourced & \cellcolor{green!25}Free-form & ROUGE-L, BLEU-4,\\
        &&&&(tweets in news articles.) &&&& METEOR\\
        
        \midrule
        &&&&Conversational QA\\
        \midrule
        CoQA\cite{Reddy2019CoQAAC} & 2019 & EN & Mixed & Mixed & Crowdsourced & Crowdsourced & \cellcolor{green!25}Free-form & F1\\
         
        ShARC\cite{Saeidi2018InterpretationON} & 2018 & EN & Rules & Government Websites & Crowdsourced & Crowdsourced & \cellcolor{purple!25}Multi. Choices \& &  Accuracy, \\
        &&&&&&&\cellcolor{green!25}Follow up question &BLEU-1,2,3,4\\
        
        QuAC\cite{Choi2018QuACQA} & 2018 & EN & Wikipedia & Wikipedia & Crowdsourced & Crowdsourced & \cellcolor{blue!25}Span of words & F1, HEQ-Q,D\\
        
        \midrule
        &&&&Multi-hop QA\\
        \midrule
        HotpotQA\cite{Yang2018HotpotQAAD} & 2018 & EN & Wikipedia & Wikipedia & Crowdsourced & Crowdsourced & \cellcolor{blue!25}Span of words & F1, EM\\
        
        2WikiMulti- & 2020 & EN & Wikipedia & Wikipedia & KB + Template & KB & \cellcolor{blue!25}Span of words & F1, EM\\
        HopQA\cite{Ho2020ConstructingAM}\\

        HybridQA\cite{Chen2020HybridQAAD} & 2020 & EN & Wikipedia & Wikipedia & Crowdsourced & Crowdsourced & \cellcolor{blue!25}Span of words & F1, EM\\
        
        \midrule
        &&&&Long-form QA\\
        \midrule
        ELI5\cite{Fan2019ELI5LF} & 2019 & EN & Web. & Common Crawl & Reddit & Reddit & \cellcolor{green!25}Free-form & ROUG-1,2,L\\
        
        MASH-QA\cite{zhu2020question} & 2020 & EN & Health & WebMD & QA section & QA section & \cellcolor{blue!25}Span of words & F1, EM\\
        
        NLQuAD\cite{Soleimani2021NLQuADAN} & 2021 & EN & News & BBC & interrogative- & body- & \cellcolor{blue!25}Span of words & IoU \\ &&&&&sub-heading & paragraphs\\
        
        \midrule
        &&&&Cross-language QA\\
        \midrule
        XQA\cite{liu-etal-2019-xqa} & 2019 & multi(9) & Wikipedia & Wikipedia & \cellcolor{yellow!25}Synthetic Cloze & \cellcolor{yellow!25}Synthetic Cloze & Fill in entity & F1, EM\\
        
        XQuAD\cite{artetxe-etal-2020-cross} & 2020 & multi(10) & Wikipedia & SQuAD1.1 & SQuAD1.1 & SQuAD1.1 & \cellcolor{blue!25}Span of words & F1\\
        
        MLQA\cite{lewis-etal-2020-mlqa} & 2020 & multi(7) & Wikipedia & Wikipedia & Crowdsourced & Crowdsourced & \cellcolor{blue!25}Span of words & F1, EM\\
        
        \midrule
        
    \end{tabular}
\end{table*}%


\begin{table*}[htbp]
    \centering
    \captionsetup{justification=centering}
    \caption{Statistics of Machine Reading Comprehension datasets.} \label{tab:MRC_stats}%
    \setlength{\tabcolsep}{1pt} 
    \renewcommand{\arraystretch}{1.5} 
    \begin{tabular}{@{\extracolsep{3pt}}*1l@{} @{\extracolsep{3pt}}*6c@{} @{\extracolsep{3pt}}*1r@{}}
        \toprule
        &  \multicolumn{3}{c}{\textbf{Size(\#Questions)}} &&&&\textbf{Ave.Length of}\\\cline{2-5}
        \textbf{Dataset} & \textbf{total} & \textbf{training} & \textbf{dev} & \textbf{testing} & \textbf{Corpus Size} & \textbf{Corpus unit} & \textbf{Context}\\
        \hline
        &&&&Classical MRC\\
        \midrule
         CNN\cite{hermann2015teaching} & 387K & 380K & 4K & 3K & 93K & Document & 742\\
        Daily Mail\cite{hermann2015teaching} & 997K & 879K & 65K & 53K & 220K & Document & 809\\
        
        CBT\cite{Hill2016TheGP} & 687K & 669K & 8K & 10K & 108 & Book & 463 \\
        
        BookTest\cite{Bajgar2016EmbracingDA} & 14.16M & 14.14M & 10K & 10K & 14K & Book & 379 \\
        
        HLF-RC\cite{cui-etal-2016-consensus} & 873K & 870K & 3K & 4.6K & 108 & Document & 379 \\
        
        \midrule
        
        MCTest\cite{richardson-etal-2013-mctest} & 2K & 1.2K & 200 & 600 & 500 & Story & 212 \\
        
        
        MovieQA\cite{Tapaswi2016MovieQAUS} & 21K & 14K & 3K & 4K & 548 & Movie & 728 \\
        
        RACE\cite{Lai2017RACELR} & 98K & 88K & 5K & 5K & 28K & Passage & 322 \\
        
        \midrule
        
        SQuAD1.1\cite{Rajpurkar2016SQuAD1Q}& 108K & 88K & 10K & 10K & 536 & Document & 117 \\
        SQuAD2.0\cite{Rajpurkar2018KnowWY}& 151K & 130K & 12K & 9K & 505 & Document & 117 \\
        
        NewsQA\cite{Trischler2017NewsQAAM} & 120K & 93K & 5K & 5K & 13K & Document & 749 \\
        
        SearchQA\cite{Dunn2017SearchQAAN} & 140K & 100K & 14K & 27K & 140K & Passage & 59 \\
        
        TriviaQA-Web\cite{Joshi2017TriviaQAAL} & 96K & 76K & 10K & 10K & 663K & Document & 2,895 \\
        
        TriviaQA-Wiki\cite{Joshi2017TriviaQAAL} & 78K & 62K & 8K & 8K & 139K & Document & - \\
        
        NaturalQS\cite{Kwiatkowski2019NaturalQA} & 323K & 307K & 8K & 8K & 323K & Document & 7,320 \\
        
        \midrule
        
        MS-MARCO\cite{nguyen2016ms} & 1M & 809K & 101K & 101K & 8.8M & Passages & 56 \\
        
        NarrativeQA\cite{Kocisk2018TheNR} & 47K & 33K & 3K & 11K & 1,572 & Document & 659/62K \\
        
        DuReader\cite{He2018DuReaderAC} & 200K & 181K & 10K & 10K & 1M & Document & 396 \\
        
        TweetQA\cite{Xiong2019TWEETQAAS} & 14K & 11K & 1K & 2K & 18K & Tweet & 32 \\
        
        \midrule
        &&&&Conversational QA\\
        \midrule
        
        CoQA\cite{Reddy2019CoQAAC} & 127K & 110K & 7K & 10K & 8K & Passage & 271 \\
                
        ShARC\cite{Saeidi2018InterpretationON} & 948 & 628 & 69 & 251 & 32K & Utterance & - \\
        
        QuAC\cite{Choi2018QuACQA} & 98K & 84K & 7K & 7K & 9K & Unique Section & 397 \\
        
        \midrule
        &&&&Multi-hop QA\\
        \midrule
        HotpotQA\cite{Yang2018HotpotQAAD} & 105K & 90K & 7K & 7K & 2/question & Passage & 917 \\
        
        2WikiMulti- & 193K & 167K & 13K & 13K & 2/question & Passage & -\\
        HopQA\cite{Ho2020ConstructingAM}\\
        
        HybridQA\cite{Chen2020HybridQAAD} & 70K & 63K & 3.5K & 3.5K & 13K & table & -\\

        \midrule
        &&&&Long-form QA\\
        \midrule
        ELI5\cite{Fan2019ELI5LF} & 272K & 237K & 10K & 25K & 15/question & passage & 857.6 \\
        
        MASH-QA\cite{zhu2020question} & 35K & 28K & 3.5K & 3.5K & 1/question & Document & 696 \\
        
        NLQuAD\cite{Soleimani2021NLQuADAN} & 31K & 25K & 3K & 3K & 13K & Document & 877\\
        
        \midrule
        &&&&Cross-language QA\\
        \midrule
        XQA\cite{liu-etal-2019-xqa} & 90K & 56K & 17K & 17K & 10/question & Document & Vary by language\\
        &&&&&&&(256 to 1159)\\
        
        XQuAD\cite{artetxe-etal-2020-cross} & 1K & - & - & - & 1K & passage & Vary by language\\
        &&&&&&&(126.5 to 232.4)\\
        
        MLQA\cite{lewis-etal-2020-mlqa} & 46K & - & 4K & 42K & 40K & passage & Vary by language\\
        &&&&&&&(102.2 to 195.1)\\
        
        \midrule

    \end{tabular}
\end{table*}%

\subsection{Classical MRC}
Next, we will introduce classical MRC benchmark datasets by annotation style.

\subsubsection{Cloze-style MRC}

\textbf{CNN/Daily Mail(DM)}\cite{hermann2015teaching} are two large-scale cloze style MRC datasets, constructed by researchers from Google DeepMind using the online news articles from CNN and Daily Mail. Summaries and paraphrase sentences, along with the associate documents are converted to context-query-answer triples using simple entity detection and anonymization algorithms.  Specifically, the Cloze style QA pairs are constructed by replacing one entity in the summarization points at a time with a placeholder. The rest of the article is taken as the context. Additionally, all named entities in the dataset are replaced by anonymous tokens. This forces a system to rely solely on context information, instead of transferring the meaning of the named entities between documents. In total, the combined corpus consists of over 1.3M instances. Table~\ref{tab:MRC_stats} gives more detailed statistics. The metric for evaluation is accuracy.\\

\noindent\textbf{CBT}\cite{Hill2016TheGP} is a large-scale cloze style MRC dataset proposed by Facebook AI Research. The data is collected from free children’s books through Project Gutenberg\footnote{\url{https://www.gutenberg.org/}}. Since the summaries are not available in the books as in the news articles, each CBT instance is constructed by using 20 consecutive sentences from the story as the context, with the subsequent sentence turning into a cloze style question. In each question, a word is replaced by a placeholder. A system is required to select an answer from a collection of 10 candidate words selected from the context sentences. There are roughly 687K instances in CBT. More detailed statistics are shown in Table~\ref{tab:MRC_stats}. The evaluation metric is accuracy.\\
  
\noindent\textbf{BookTest}\cite{Bajgar2016EmbracingDA} is a large-scale cloze-style MRC dataset proposed by researchers from IBM Watson. Like the CBT dataset, BookTest is derived from Project Gutenberg books using the same procedure. Because the BookTest datasets uses more books than the CBT dataset it is over 60 times larger. In total, it contains over 14M training examples and 7.9B tokens. See Table~\ref{tab:MRC_stats} for more detailed statistics. Like CBT, accuracy is used as the evaluation metric.\\
  
\noindent\textbf{HLF-RC}\cite{cui-etal-2016-consensus} is a Chinese large-scale cloze-style MRC dataset proposed by iFLYTEK Research in China. It is constructed from two data sources, People Daily, and Children’s Fairy Tale(CFT). The instances from the CFT are solely used for out-of-domain evaluation purposes. The construction method of HLF-RC is similar to the CBT and BookTest datasets, however instead of using the 20 consecutive sentences before the question sentence as the context, it uses the whole document as the context of an instance, and it only selects nouns as answers. There are about 873K instances included in this dataset. See Table~\ref{tab:MRC_stats} for more details. A system is required to predict the answer word from the whole vocabulary given the context document and the cloze-style question. The performance is evaluated by accuracy.\\
 

\subsubsection{Multi-choice MRC}

\noindent\textbf{MCTest}\cite{richardson-etal-2013-mctest} is one of the earliest benchmark datasets that was widely used to evaluate machine learning-based MRC models. It is a relatively small dataset proposed by researchers from Microsoft Research in 2013. Two thousand multiple choice questions based on 500 passages were collected through Amazon Mechanical Turk.  The turkers were asked to write a short fictional children's story for any topic. Four multiple choice questions were created based on each story. Each question was paired with one golden answer and three reasonable distractor choices. Detailed dataset statistics are given in Table~\ref{tab:MRC_stats}. Systems are evaluated by their question answering accuracy given the associated story.\\
    
    
\noindent\textbf{MovieQA}\cite{Tapaswi2016MovieQAUS} is an unusual MRC dataset because it uses text passages and video clips for context. Systems need to understand the complex semantics underlying different forms of data, which requires the use of different semantic extraction models within the system. Hence, such tasks are also called multi-model MRC. The MovieQA dataset consists of 15K multi-choice questions based on 408 movie plots. Each question has 5 deceiving choices of which one is correct. See Tabel~\ref{tab:MRC_stats} for statistics. Multiple information sources are included in the context of each instance, including plots, subtitles, video clips, scripts, and DVS transcriptions\cite{Rohrbach2015ADF}. Accuracy is used for evaluation.\\
  
\noindent\textbf{RACE}\cite{Lai2017RACELR} is a large-scale multi-choice MRC dataset proposed by researchers from Carnegie Mellon University. It is derived from the English exams for Chinese middle and high school students. RACE contains 98K multi-choice questions based on 28K passages where each question consists of four candidates of which one is correct. See Table~\ref{tab:MRC_stats} for more details. The exams were designed by experts to evaluate students’ English reading comprehension. Exam questions cover a broad range of topics and include complex questions requiring passage-wise reasoning. Questions in RACE are further categorized by their difficulty, i.e. the middle school level and the high school level. Systems are evaluated by their question answering accuracy given the associated passage.


\subsubsection{Span Extraction MRC}
\textbf{SQuAD} has two versions: SQuAD 1.1\cite{Rajpurkar2016SQuAD1Q} and SQuAD 2.0\cite{Rajpurkar2018KnowWY}. SQuAD 1.1 is a large-scale span extraction style MRC dataset proposed by researchers from Stanford University in 2016. It contains about 108K crowdsourced QA pairs with associated passages extracted from Wikipedia. The annotators were required to ask up to five questions based on each given context passage and select a segment of text from the passage as the answer for every question. Every question in this dataset is guaranteed to have an answer. \\
  
SQuAD 2.0 extends the SQuAD 1.1 by adding roughly 54K unanswerable questions created by crowd workers. This time, each annotator was asked to create up to five unanswerable questions for each given context passage. Table~\ref{tab:MRC_stats} gives detailed statistics for both SQuAD 1.1 and SQuAD 2.0 datasets. A system tested on SQuAD 2.0 is required to fulfill two tasks. First, it has to decide whether a question is answerable given a context passage. Then, if the question is answerable, it needs to select a span of words from the context passage as the answer to the question. The evaluation metrics for the second task are exact match(EM) and F1.\\
  
\noindent\textbf{NewsQA}\cite{Trischler2017NewsQAAM} is a large-scale extractive MRC dataset proposed by researchers from Microsoft. It consists of about 120K crowdsourced questions based on 13K news articles from CNN. Unlike SQuAD, NewsQA was annotated by two sets of crowd workers. One set of annotators only had access to the highlights of the articles. Using the highlights, they created up to three questions about the article. The other set of annotators were required to answer these questions using spans of words from the full articles. Such an annotation method mimics a user's realistic answer searching activity and prevents trivial lexical matching between questions and answers. It also gives unanswerable questions for the context articles. Statistical details are presented in Table~\ref{tab:MRC_stats}. The evaluation metrics are EM and F1.\\
  
\noindent\textbf{SearchQA}\cite{Dunn2017SearchQAAN} is a large-scale extractive MRC dataset proposed by researchers from the New York University. It is intended to reflect the general process of question-answering. Unlike SQuAD and NewsQA, SearchQA starts by collecting QA pairs scraped from J!Archive, then text snippets are retrieved by Google using each question as the search query, making sure the answers are contained in the snippets and the snippets cannot be found easily by matching the words in the questions. This results in over 140K QA pairs, in which each QA pair has on average of 49.6 snippets as context. Detailed statistics are shown in Table~\ref{tab:MRC_stats}. Evaluation is done by accuracy and F1 measure.\\
  
\noindent\textbf{TriviaQA}\cite{Joshi2017TriviaQAAL} is a large-scale extractive MRC dataset proposed by researchers from the Allen Institute for Artificial Intelligence(AI2). Similar to the SearchQA dataset, it first collects QA pairs from QA websites, then collects relevant documents using a search engine, making sure the answer spans are in the documents. Specifically, in TriviaQA, 96K unique QA pairs are collected from 14 trivia and quiz-league websites. Documents are collected using the Bing API from two sources: web documents and Wikipedia articles. Filters are applied to remove questions that are shorter than 4 tokens and documents that do not contain the correct answer string. Finally, this dataset is released in three versions: Trivia-Web uses the top 10 web documents returned by the Bing API as the context for each question. Trivia-wiki uses the content of all the Wikipedia pages that contain the entities in the question as context. Trivia-unfiltered contains 110K QA pairs and 740K documents. More details are provided in Table~\ref{tab:MRC_stats}. Evaluations are done using EM and F1 measures.\\
  
\noindent\textbf{Natural Questions}\cite{Kwiatkowski2019NaturalQA} is a large-scale extractive MRC dataset proposed by Google Research. It contains over 323K instances in the form of tuples composed of a question, a related Wikipedia article, a long answer, and a short answer. The questions are from Google user search logs. Each answer is annotated in long and short forms. The answer is based on a Wikipedia article from the top five search results.  A long answer could be in the form of a table, a truncated list, or the entire list from an article, but it typically is a paragraph. A short answer consists of an entity or a set of entities within the long answer. For binary questions, a short answer can be "yes" or "no". "No answer" is an allowed flag for both long and short answers. For detailed statistics see Table~\ref{tab:MRC_stats}. For both long and short answer extraction tasks, the evaluation metric is F1 measure.
  

\subsubsection{Free-form MRC}
\textbf{MS-MARCO}\cite{nguyen2016ms} is a large-scale free-form MRC dataset proposed by Microsoft AI \& Research. It contains over 1M questions sampled from the search query logs of Bing or Cortana. Each question may have zero, one, or multiple human-generated answers. Each answer is manually created based on passages from the top retrieved documents. During the annotation process, the passages are ranked by their relevance to the question, and the answer is refined continuously for higher accuracy and quality. On average, each question is paired with ten relevant passages as the context. Table~\ref{tab:MRC_stats} gives the statistics of this dataset. This dataset supports three benchmark tasks: (1)Rerank 1000 candidate passages retrieved by BM25 by their relevance to the question; (2)Retrieve the most relevant passages from the entire corpus; (3)Given a question and the corresponding context passages, generate the correct answer when the question is answerable, otherwise, return “No Answer Present”. For the first two passage retrieval tasks, the evaluation metric is MRR@10; for the last MRC task, the evaluation metrics are ROUGE-L and BLEU-1 measures.\\
  
\noindent\textbf{NarrativeQA}\cite{Kocisk2018TheNR} is a large-scale free-form MRC dataset proposed by researchers from Google Deep Mind. It contains 47K instances of human-written questions and answers based on human-generated abstractive summaries of books and movie scripts. Books used are from Project Gutenberg and movie scripts are collected from the internet. All context documents are collected as stories in this dataset. The stories are matched with plot summaries from Wikipedia using titles and verified by human annotators. This results in about 1.6K stories and corresponding summaries. Thirty QA pairs are then crowdsourced for each summary. Annotating based on summaries isolates the annotators from the full stories, which prevents creating answers based on localized context. Answers are allowed to be of various lengths, from one word to a few sentences. All sentences had to be grammatically correct. Table~\ref{tab:MRC_stats} gives detailed statistics. A system is required to generate the answer given the question and the associated full story. The performance is measured by ROUGE-L and BLEU-1.\\
  
\noindent\textbf{DuReader}\cite{He2018DuReaderAC} DuReader is a Chinese large-scale free-form MRC dataset proposed by researchers from Baidu in China.  Similar to MS MARCO, DuReader consists of sampled questions from real users’ search queries. Questions are from Baidu Search (a Chinese search engine) or Baidu Zhidao (a Chinese community QA website).  The questions were sampled through a multistep process: first, the most frequent 1M  queries were sampled; second, three different question types (Entity, Description, and YesNo) were selected using a pre-trained classifier (with over 90\% recall); finally, the remaining questions were verified and filtered by crowd workers.  This results in 200K unique questions. Then the questions were randomly split into two sets, each of which was used to retrieve context documents from the two sources respectively.  A human-written answer was then created by summarizing the top 5 documents retrieved for each question. It is worth noting that, in DuReader, both “no answer” and multiple different answers are allowed. For those questions that have multiple answers, an opinion label (Yes, No, Depends) is added to each answer. Finally, all answers were further checked and improved by 52 experts. Detailed statistics are in Table~\ref{tab:MRC_stats}. A system is required to generate an answer to a question that resembles a human-generated answer. The performance is evaluated by ROUGE-L and BLERU-4 measures.\\
  
\noindent\textbf{TweetQA}\cite{Xiong2019TWEETQAAS} is a free-form MRC dataset proposed by researchers from UCSB and IBM Research. It is also the first large-scale dataset for QA over social media data. TweetQA consists of nearly 14K (tweet, question, answer) triples, where the tweets are collected from news websites (CNN, NBC). QA pairs are manually created by crowd workers based on the tweets. In general, there are 18K tweets extracted from 11K news articles.  Annotators were asked to create answers using words that were not in the context tweet. Binary questions and questions shorter than 5 words were removed. Detailed information is listed in Table~\ref{tab:MRC_stats}.  TweetQA requires a system to generate a textual answer given a question and the associated tweet. The performance is evaluated by ROUGE-L, BLEU-1, and METEOR.
  
\subsection{Novel MRC}
Next, we will introduce MRC datasets in four novel application scenarios: conversation question answering, multi-hop question answering, long-form question answering, and cross-language question answering. We refer to these tasks as novel MRC.

\subsubsection{Conversational Question Answering}
\begin{figure}
    \centering
    \includegraphics[width=\linewidth]{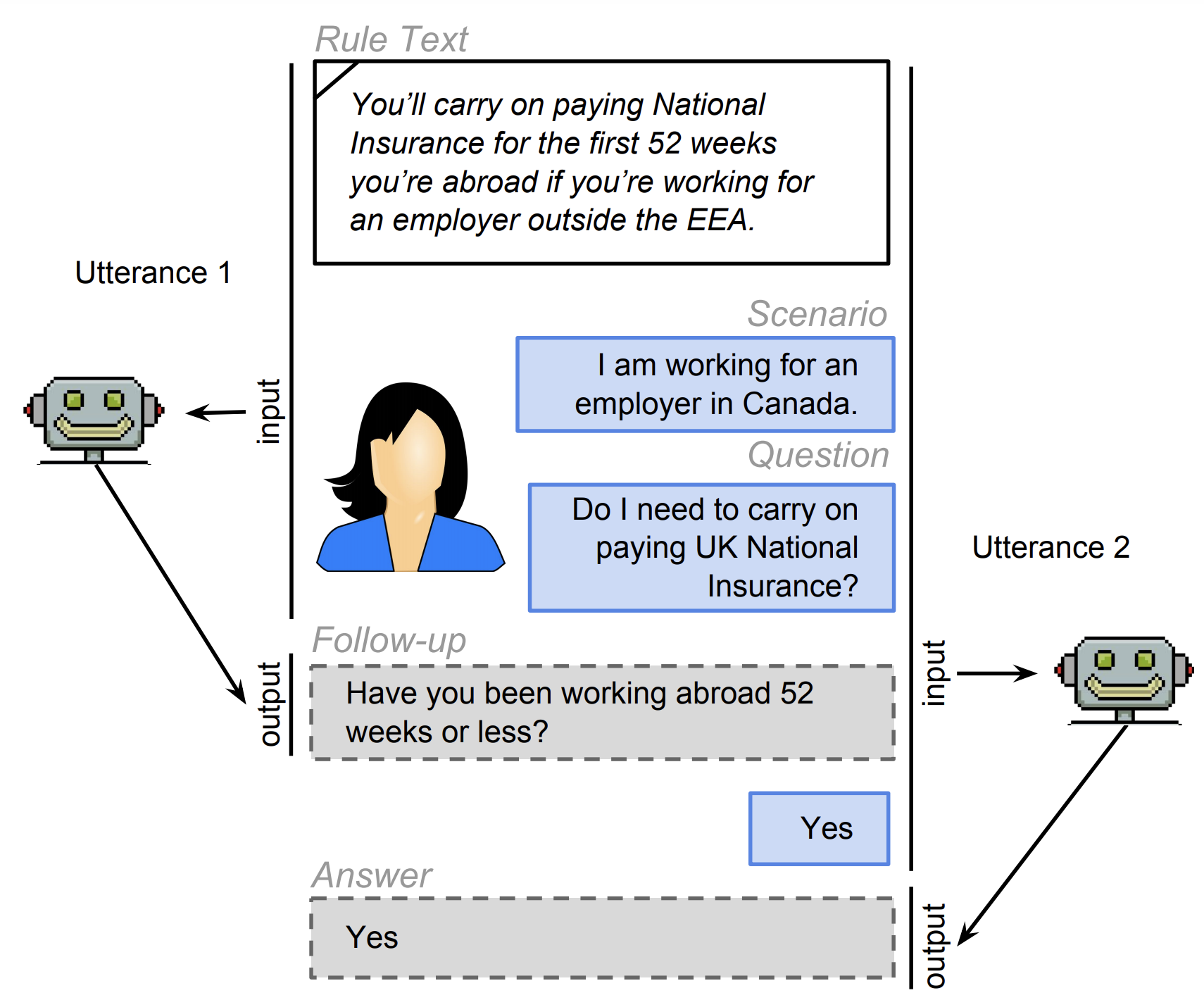}
    \caption{An example of the conversational QA from ShARC\cite{Saeidi2018InterpretationON}.}
    \label{fig:ShARC}
\end{figure}

Conversational Question Answering (CQA) is a task developed from classical MRC tasks. The major difference between classical MRC and CQA is the form of the context information used to answer the question. Specifically, CQA requires a system to understand a context passage and understand the conversation history to engage in multi-turn QA. When appropriate, the "answer" could even be a follow-up question. Figure~\ref{fig:ShARC} gives an example of such a requirement. CQA extends the classical MRC application scenario from a single turn QA to a multi-turn dialog style QA.\\

\textbf{CoQA}\cite{Reddy2019CoQAAC} is the first public MRC dataset designed for conversational QA. It requires a system to answer a series of questions in a conversation based on a context that consists of a text passage and conversation history before the question. In this dataset, to reflect the nature of human conversations, all but the first question are designed to depend on the conversation history. The answers can be free-form text while a text span from the passage is also provided for each answer as a rationale. Context passages across multiple domains are used to construct conversations. Specifically, 8K conversations with on average 15 turns are constructed based on passages from seven different domains. The evaluation metric for this dataset is F1.\\

\textbf{ShARC}\cite{Saeidi2018InterpretationON} is designed to tackle a challenging situation in MRC where the answer is not directly expressed in the context passage of rules because the questions are underspecified. In this case, systems have to ask clarification questions to better understand the users’ condition or situation. This dataset consists of 32k instances based on real-world rules collected from government websites and crowdsourced questions and scenarios. Figure~\ref{fig:ShARC}, borrowed from the original paper\cite{Saeidi2018InterpretationON}, gives an example of two utterances for the rule interpretation. Specifically, this dataset consists of two stages of tasks: multiple-choice (i.e., Yes, No, Irrelevant, More) and follow-up question generation. For the multiple-choice task, given the conditions (the question, the context passage, and the scenario), a classifier model has to answer "Yes", "No", "Irrelevant" or "More". “Yes” and “No” are answers to the question, “Irrelevant” means that the question is irrelevant to the context passage, and “More” means a follow-up question is needed. For the follow-up question generation task, when the early-stage task predicts “More”, a question-generation model needs to generate a follow-up question. The metrics for the multiple-choice task are micro and macro accuracy. The metrics for the follow-up question generation task are BLEU-1,2,3,4.\\

\textbf{QuAC}(Question Answering in Context)\cite{Choi2018QuACQA} is a large-scale conversational QA dataset that emphasizes context and open-endedness. In QuAC 86\% of questions require contextual information (e.g, conversation history, context article, etc.), and over half of questions are non-factoid questions annotated with multiple reference answers. To create a dialogue crowd workers were placed in a teacher-student scenario. In each dialog, the student asks questions about an entity, \textit{e}, given only the sections’ title, \textit{t}, and the first paragraph, \textit{b}, of a Wikipedia article. The teacher has access to the full section text and answers the questions with spans from the text. Not allowing the student access to the evidence text reduces the lexical overlapping between questions and answers. Fourteen thousand dialogues were collected in total. This dataset requires a system to predict the answer span indices \textit{i}, \textit{j} in the section text, \textit{s}, given the conversation history and the supporting material including the entity, \textit{e}, topic, \textit{t}, background, \textit{b}, and section text, \textit{s}. The evaluation metrics are HEQQ, HEQD, and F1.
\\


\subsubsection{Multi-hop Question Answering}
\begin{figure}
    \centering
    \includegraphics[width=\linewidth]{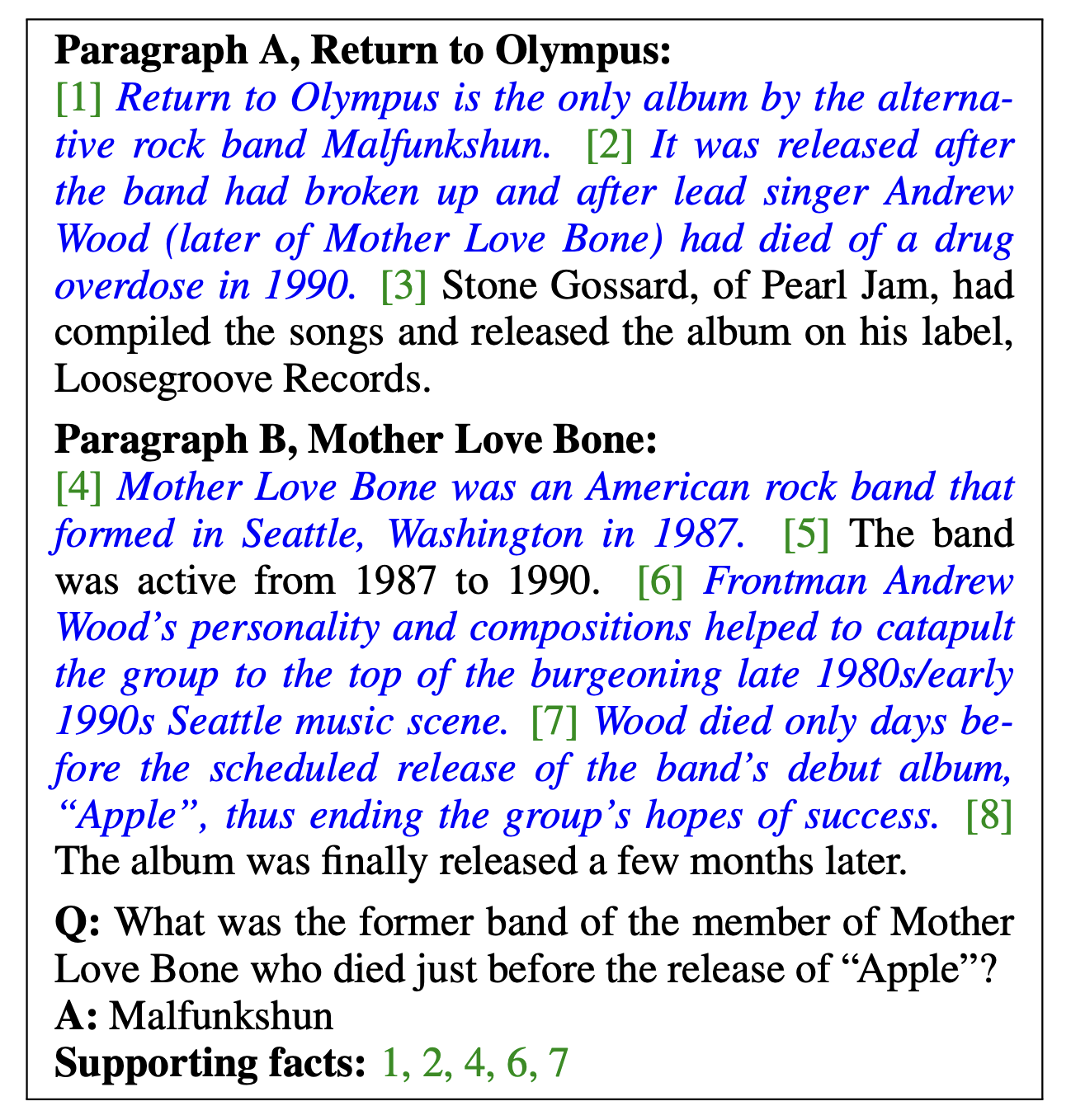}
    \caption{An example of the multi-hop QA from HotpotQA\cite{Yang2018HotpotQAAD}.}
    \label{fig:Hotpot}
\end{figure}

Unlike the classical MRC tasks that focus on answering questions that only require single-step reasoning, multi-hop question answering focuses on answering complex questions that require multi-step reasoning.  Multi-step reasoning usually involves using multiple documents. Figure~\ref{fig:Hotpot} illustrates an example of the multi-hop QA that requires two-step reasoning over two paragraphs.\\

\textbf{HotpotQA}\cite{Yang2018HotpotQAAD} contains 113K question-answer pairs. Each QA pair is constructed using a pair of Wikipedia documents to make sure multi-step reasoning over both of the documents is necessary when answering the question. Answers are annotated using the spans of text from the context. Additionally, supporting fact sentences are annotated to improve system explainability. Figure~\ref{fig:Hotpot}, borrowed from \cite{Yang2018HotpotQAAD}, gives an example of the multi-hop QA in HotpotQA. There are two benchmark settings provided by this dataset, the distractor setting, and the full-wiki setting. In the distractor setting, the context of each QA pair consists of ten paragraphs, two of which are the gold paragraphs used to construct the QA pair and the rest are the top-eight paragraphs retrieved from Wikipedia using bigram tf-idf\cite{chen-etal-2017-reading}. In the full-wiki setting, systems are challenged to answer the question given in the first paragraphs of all Wikipedia documents, which is also known as machine reading at scale (MRS) or open-domain question answering(ODQA), which is introduced in the section~\ref{sec:ODQA}. The metrics for both settings are EM and F1.\\

\textbf{2WikiMultiHopQA}\cite{Ho2020ConstructingAM} is a large-scale multi-hop dataset that contains 193K natural questions of four types: comparison, inference, compositional, and bridge comparison.  The answers are also spans of words within the context passages. The term “2Wiki” means it uses both Wikidata (structured data) and  Wikipedia articles (unstructured data) to generate instances. Each instance consists of a QA pair, two Wikipedia summary passages, supporting sentences, and evidence.  Evidence consists of triplets in the Wikidata indicating the inference path of the answer, which is a major difference from HotpotQA.  The questions are generated programmatically from 28 predefined templates where logical rules are applied to guarantee multi-step inference. This dataset can be used for three subtasks: (1) given question, Q, predict a span in a set of documents, D, as an answer, (2) provide a set of supporting sentences from D, and (3) give a reasoning path in triplets. EM and F1 are the metrics for all subtasks.\\

\textbf{HybridQA}\cite{Chen2020HybridQAAD}, unlike other multi-hop QA datasets that aim at multi-step reasoning over multiple documents\cite{Yang2018HotpotQAAD} \cite{Ho2020ConstructingAM}, requires a system to reason over heterogeneous information: tabular and textual data. This dataset simulates a setting where the evidence is distributed among heterogeneous data, and the system has to answer a question by aggregating information from different forms.  The context of each question consists of a Wikipedia table and multiple documents linked with the entities in the table. The questions are designed to be unanswerable using a single form of data. The answers are minimum text spans from either a table cell or a specific passage. Nearly 70K questions are annotated by crowd workers and checked by experts.  Given a table and associated passages, the task is to answer a multi-hop question by predicting a span of words from either the table or the passages. The metrics are EM and F1.


\subsubsection{Long-form Question Answering}
\begin{figure}
    \centering
    \includegraphics[width=\linewidth]{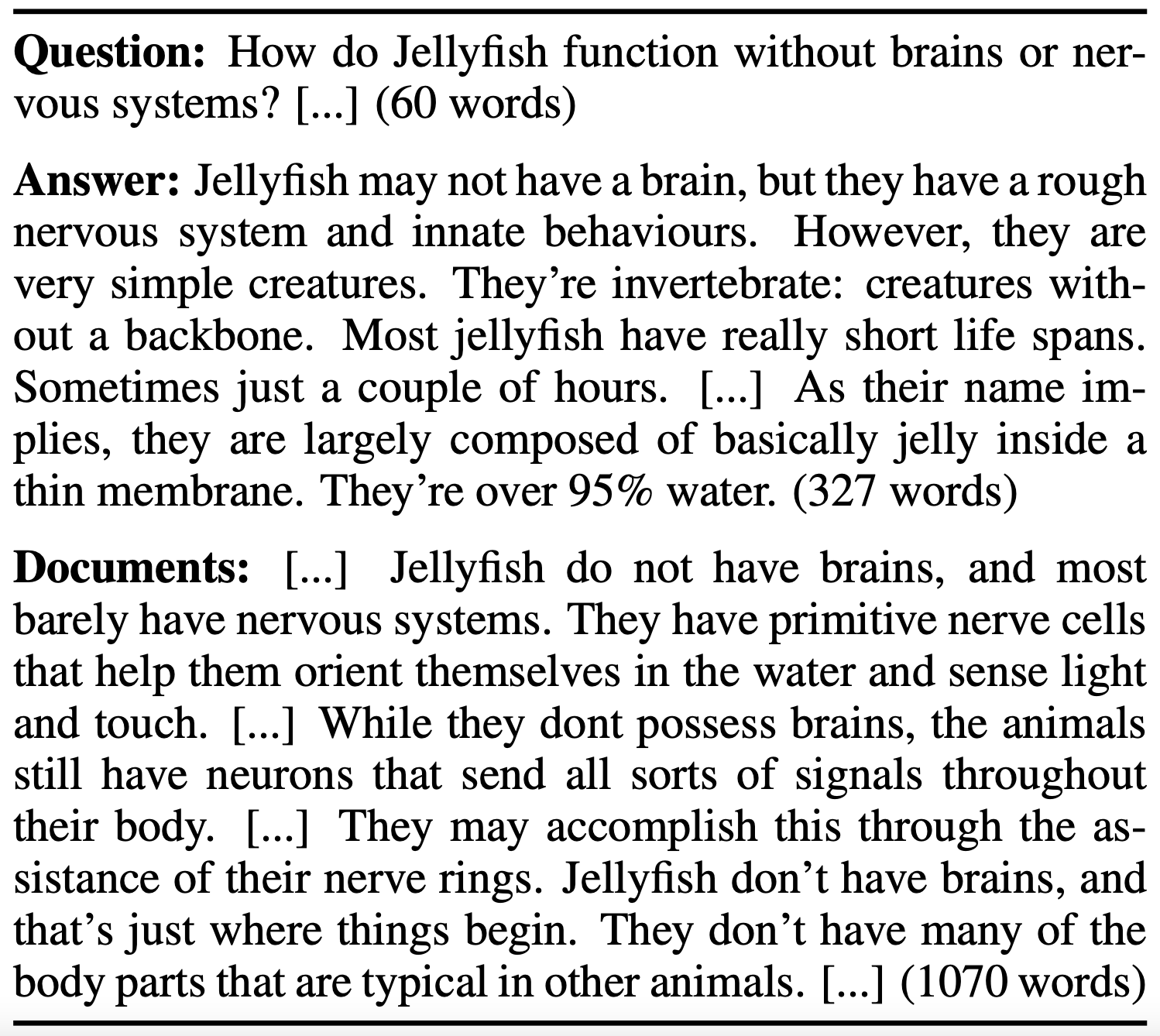}
    \caption{An example of the long-form QA from ELI5\cite{Fan2019ELI5LF}.}
    \label{fig:ELI5}
\end{figure}

Classical MRC datasets usually have concise, unambiguous answers, but can be problematic when answering complex questions requiring more explanation, e.g., open-ended questions.  Long-form QA solves this problem by providing long-form answers across multiple sentences. Figure~\ref{fig:ELI5} gives an example of long-form QA from the ELI5 dataset where the question is not answerable with a short response.\\

\textbf{ELI5}\cite{Fan2019ELI5LF} is the first large-scale long-form QA dataset. It contains 272K open-ended questions and top-voted answers that are collected from the Reddit forum “Explain Like I’m 5”.  The answers are provided by an online community, and they are supposed to be understandable to five-year-olds. The average length of the answers is 130.6 words. The context of each question is collected in two steps. First, the top 100 best matching web documents are retrieved from the Common Crowl\footnote{\url{http://commoncrawl.org}}. Then the sentences are extracted and concatenated based on their tf-idf score to the question to serve as the context of the question.  Figure~\ref{fig:ELI5}, borrowed from\cite{Fan2019ELI5LF}, illustrates an example of ELI5. This dataset requires a system to generate a paragraph-length answer to a complex question given the supporting context. The evaluation metrics are ROUGE-1, ROUGE-2, and ROUGE-L.\\

\textbf{MASH-QA}\cite{zhu2020question} is a large-scale span extraction (sentence selection) style MRC dataset in the Healthcare domain. It provides 35K questions with long-form answers that consist of multiple nonconsecutive sentences spanning across a long document. The average length of answers is 67.2 words. The questions and answers are collected from the health website WebMD, which contains consumer healthcare-related articles on different topics and common healthcare problems and corresponding answers under each topic. The sentences of each answer are mapped to the corresponding sentences in the relevant article using tf-idf. This dataset requires a system to select all the annotated sentences from a context article as a long-form answer given a question. Precision, Recall, and F1 measures are used for evaluation.\\

\textbf{NLQuAD}\cite{Soleimani2021NLQuADAN} is a large-scale span extraction style MRC dataset that contains 31K non-factoid questions and long answer spans from 13K BBC news articles. All the annotations in NLQuAD are created programmatically by extracting interrogative sub-headings in the articles as questions and body paragraphs corresponding to the sub-heading as contiguous answers to the questions. Human evaluations are done to guarantee the quality of the answers. The average length of the answers is 175 words. Unlike other span extraction style datasets, NLQuAD uses intersection over union (IoU) for system evaluation.


\subsubsection{Cross-language Question Answering}
Large scale MRC benchmark datasets boosted the deep learning-based language models which achieved remarkable results in Textual QA tasks. But one major problem is that most of them are in English. It is intractable and unrealistic to collect and label large-size training data for every language, especially for those low-resource languages. Cross-language QA is proposed to tackle this problem by learning language-independent features that could be used to model the answer prediction probability P(A $|$ Q, C) for a target language under the supervision of the source language. Here A, Q, C stand for the answer, question and context respectively.\\

\textbf{XQA}\cite{liu-etal-2019-xqa} is a cross-language cloze style QA dataset where QA pairs are collected from the “Did you know” section on the main page of Wikipedia.  The “Did you know” section contains factual questions from human editors in multiple languages. Each entity in the question has a link to its corresponding Wikipedia page. The questions in the XQA dataset are constructed by taking each “Did you know” question from the section and masking one entity. The entity name and its aliases from the Wikidata knowledge base are used as golden answers. For each question, the top-10 Wikipedia articles retrieved by a BM25 retriever are collected as the context. It is worth noting that, the first paragraph of each document in the context is removed to avoid trivial predictions. In total, 62K questions are annotated for English and 28K for eight other languages. Table~\ref{tab:XQA_Stats} shows the detailed statistics of the XQA dataset. EM and F1 are the metrics used for evaluation.\\ 

\begin{table}[t]
    \centering
    \captionsetup{justification=centering}
    \caption{Statistics of the XQA datasets.}
    \label{tab:XQA_Stats}
    \setlength{\tabcolsep}{3pt} 
    \renewcommand{\arraystretch}{1.5} 
    \begin{tabular}{@{\extracolsep{11pt}}*1l@{} @{\extracolsep{11pt}}*3c@{} @{\extracolsep{11pt}}*1r@{}}
        \toprule
        \textbf{Language} & \textbf{Train} & \textbf{Dev} & \textbf{Test} & \textbf{Ave.Context Length}\\
        \hline
        \midrule
        English & 56,279 & 2,926 & 2,924 & 735.91\\
        Chinese & - & 2,532 & 2,535 & 1159.28\\
        French & - & 1,946 & 1,749 & 913.72\\
        German & - & 3,895 & 3,804 & 450.65\\
        Polish & - & 924 & 922 & 256.78\\
        Portuguese & - & 359 & 348 & 482.74\\
        Russian & - & 3,590 & 3,490 & 503.28\\
        Tamil & - & 597 & 586 & 200.45\\
        Ukrainian & - & 589 & 615 & 584.93\\
        \midrule

    \end{tabular}
\end{table}%

\textbf{XQuAD}\cite{artetxe-etal-2020-cross} is a span extraction style cross-language MRC dataset proposed to evaluate the generalization ability of deep multilingual language models. It consists of 240 paragraphs and 1190 QA pairs that are translated from the development set of the SQuAD1.1\cite{Rajpurkar2016SQuAD1Q} dataset into ten languages (Spanish, German, Greek, Russian, Turkish, Arabic, Vietnamese, Thai, Chinese, and Hindi) by professional translators through the Gengo\footnote{\url{https://gengo.com}} translation service. F1 score is used for evaluation.\\

\textbf{MLQA}\cite{lewis-etal-2020-mlqa} is a multi-way aligned cross-language MRC evaluation dataset annotated in a span extraction style. It contains QA instances in 7 languages: English, Arabic, German, Spanish, Hindi, Vietnamese, and Simplified Chinese. In this dataset, there are over 12K instances in English and 5K instances each of the remaining languages, with each instance having on average of 4 parallels in other languages. Table~\ref{tab:MLQA_Stats} shows the detailed statistics of the MLQA dataset. To construct the MLQA dataset, first, the context passages programmatically extracted from Wikipedia articles, making sure they have aligned sentences that have the same or similar meaning in multiple languages. Then the questions are crowdsourced on the extracted English paragraphs, making sure the answer is in the aligned sentence.  Finally, the questions are translated to target languages by professional translators, and answer spans are annotated in the aligned contexts for the target languages. Such alignment reduces the cost of annotation by leveraging naturally-written documents and avoiding passage translation. Two benchmark tasks are provided by the MLQA dataset, the first task, \textit{cross-lingual transfer} (XLT), requires a system trained in a source language \textit{x} (e.g. English) to extract an answer $a_y$ in language \textit{y} given context, $c_y$, and question, $q_y$, at test time; the second task, \textit{generalized cross-lingual transfer}(G-XLT), requires a system trained in a source language \textit{x} to extract $a_z$ from $c_z$ in language \textit{z} given $q_y$ in language \textit{y}. Performance is evaluated by EM and F1.

\begin{table}[t]
    \centering
    \captionsetup{justification=centering}
    \caption{Statistics of the MLQA datasets.}
    \label{tab:MLQA_Stats}
    \setlength{\tabcolsep}{3pt} 
    \renewcommand{\arraystretch}{1.5} 
    \begin{tabular}{@{\extracolsep{11pt}}*1l@{} @{\extracolsep{11pt}}*3c@{} @{\extracolsep{11pt}}*1r@{}}
        \toprule
        \textbf{Language} & \textbf{Train} & \textbf{Dev} & \textbf{Test} & \textbf{Ave.Context Length}\\
        \hline
        \midrule
        English & - & 1,148 & 11,590 & 157.5\\
        German & - & 512 & 45,17 & 102.2\\
        Spanish & - & 500 & 5,253 & 103.4\\
        Arabic & - & 517 & 5,335 & 116.8\\
        Chinese & - & 504 & 5,137 & 222.9\\
        Vietnamese & - & 511 & 5,495 & 195.1\\
        Hindi & - & 507 & 4,918 & 141.5\\
        \midrule

    \end{tabular}
\end{table}%
\section{Open-domain Question Answering}\label{sec:ODQA}

One major difference between Open-domain Question Answering(ODQA) and MRC is the scale of context information. 
In the MRC setting, a short specific context that contains the answer to the question is provided, whereas, in ODQA, systems are required to find an answer from a large document set, e.g., Wikipedia.
With the development of MRC techniques and the emergence of large-scale MRC datasets, modern neural network-based ODQA models and their corresponding benchmarks are now widely used.

Theoretically, because of the end-to-end nature of ODQA, any existing QA dataset that contains objective question-answer pairs can be curated to serve in this scenario by pairing it with a large external knowledge source that can be used to answer its questions, e.g., Wikipedia.
Hence, besides those datasets that were originally designed for ODQA (e.g., CuratedTREC\cite{Baudis2015ModelingOT}, Quasar-T\cite{Dhingra2017QuasarDF}), many famous large-scale MRC datasets(e.g., SQuAD\cite{Rajpurkar2016SQuAD1Q}, TRIVIA QA\cite{Joshi2017TriviaQAAL}, Natural Questions\cite{Kwiatkowski2019NaturalQA}, etc.) were curated to serve as benchmarks for ODQA systems(e.g., DrQA\cite{chen-etal-2017-reading}, ORQA\cite{Lee2019LatentRF}, etc.).
Next, we will introduce the most widely-used benchmarks in modern ODQA systems. The information on these datasets is summarized in Table~\ref{ODQA_table}.

\begin{table*}[htbp]
    \centering
    \captionsetup{justification=centering}
    \caption{Open-domain question answering benchmarks.} \label{ODQA_table}%
    \setlength{\tabcolsep}{6pt} 
    \renewcommand{\arraystretch}{1.5} 
    \begin{tabular}{@{\extracolsep{6pt}}*1l@{} @{\extracolsep{6pt}}*5c@{} @{\extracolsep{6pt}}*1r@{}}
        \toprule
        \textbf{Dataset} & \textbf{Language} & \textbf{\#Questions} & \textbf{External Knowledge} & \textbf{Metrics} & \textbf{Used by}\\
        \hline
       
        CuratedTREC\cite{Baudis2015ModelingOT} & EN & 2K & Wikipedia & EM, F1 & \cite{chen-etal-2017-reading}\cite{Wang2018R3RR}\cite{Lee2019LatentRF}\cite{Guu2020REALMRL}\cite{karpukhin-etal-2020-dense}\cite{Lewis2020RetrievalAugmentedGF}\cite{roberts-etal-2020-much}, etc.\\
        
        WebQuestions\cite{Berant2013SemanticPO} & EN & 4K & Wikipedia & EM, F1 &  \cite{chen-etal-2017-reading}\cite{Wang2018R3RR}\cite{Lee2019LatentRF}\cite{Guu2020REALMRL}\cite{karpukhin-etal-2020-dense}\cite{Lewis2020RetrievalAugmentedGF}\cite{Brown2020LanguageMA}\cite{roberts-etal-2020-much}, etc.\\
        
        WikiMovies\cite{Miller2016KeyValueMN} & EN & 96K & Wikipedia & EM, F1 &  \cite{chen-etal-2017-reading}\cite{Wang2018R3RR}, etc.\\
        
        OpenSQuAD\cite{Rajpurkar2016SQuAD1Q}\cite{chen-etal-2017-reading} & EN & 97K & Wikipedia & EM, F1 &  \cite{chen-etal-2017-reading}\cite{Wang2018R3RR}\cite{seo-etal-2019-real}\cite{wang-etal-2019-multi}\cite{Lee2019LatentRF}\cite{karpukhin-etal-2020-dense}\cite{izacard-grave-2021-leveraging}, etc.\\
        
        Qusar-T\cite{Dhingra2017QuasarDF} & EN & 43K & Wikipedia & EM, F1 &  \cite{Wang2018R3RR}\cite{wang-etal-2019-multi}, etc.\\
        
        OpenTriviaQA\cite{Joshi2017TriviaQAAL}\cite{Lee2019LatentRF} & EN & 96K & Wikipedia & EM, F1 & \cite{Lee2019LatentRF}\cite{wang-etal-2019-multi}\cite{karpukhin-etal-2020-dense}\cite{Lewis2020RetrievalAugmentedGF}\cite{Brown2020LanguageMA}\cite{roberts-etal-2020-much}\cite{izacard-grave-2021-leveraging}\\
        
        OpenNaturalQuestions\cite{Kwiatkowski2019NaturalQA}\cite{Lee2019LatentRF} & EN & 315K & Wikipedia & EM, F1 & \cite{Lee2019LatentRF}\cite{wang-etal-2019-multi}\cite{karpukhin-etal-2020-dense}\cite{Lewis2020RetrievalAugmentedGF}\cite{Brown2020LanguageMA}\cite{roberts-etal-2020-much}\cite{izacard-grave-2021-leveraging}\\
        
        \midrule
        
        KILT-NaturalQuestions\cite{Kwiatkowski2019NaturalQA}\cite{Petroni2021KILTAB} & EN & 92K & Wikipedia & EM, F1, ROUGE-L & \href{https://eval.ai/web/challenges/challenge-page/689/leaderboard/1905}{KILT-NaturalQuestions Leader board}\\
        
        KILT-HotpotQA\cite{Yang2018HotpotQAAD}\cite{Petroni2021KILTAB} & EN & 100K & Wikipedia & EM, F1, ROUGE-L & \href{https://eval.ai/web/challenges/challenge-page/689/leaderboard/1906}{KILT-HOTPOT Leader board}\\
        
        KILT-TriviaQA\cite{Joshi2017TriviaQAAL}\cite{Petroni2021KILTAB} & EN & 74K & Wikipedia & EM, F1, ROUGE-L &  \href{https://eval.ai/web/challenges/challenge-page/689/leaderboard/1907#leaderboardrank-1}{KILT-TriviaQA Leader board} \\
        
        KILT-EMLI5\cite{Fan2019ELI5LF}\cite{Petroni2021KILTAB} & EN & 275K & Wikipedia & EM, F1, ROUGE-L & \href{https://eval.ai/web/challenges/challenge-page/689/leaderboard/1908}{KILT-EMLI5 Leader board}\\
        
        \midrule

    \end{tabular}
\end{table*}%

\subsection{CuratedTREC}
CuratedTREC\cite{Baudis2015ModelingOT}) is extended from the TREC QA corpus\cite{Voorhees2000BuildingAQ} which was created for ODQA.
It contains 2180 question-answer pairs from TREC 1999-2002.
In DrQA\cite{chen-etal-2017-reading}, using English Wikipedia as the external knowledge source,  CuratedTREC is used as an ODQA benchmark.
In this format, CuratedTREC is used as a benchmark in a wide range of ODQA systems, e,g., R3\cite{Wang2018R3RR}, ORQA\cite{Lee2019LatentRF}, REALM\cite{Guu2020REALMRL}, DPR\cite{karpukhin-etal-2020-dense}, RAG\cite{Lewis2020RetrievalAugmentedGF}, T5\cite{roberts-etal-2020-much}, etc.

\subsection{WebQuestions}
WebQuestions\cite{Berant2013SemanticPO} was originally designed for Knowledge base QA. It has 4K questions from the Google Suggest API and crowdsourced answers restricted to Freebase entities.
In DrQA\cite{chen-etal-2017-reading}, it was curated and used as an ODQA benchmark by converting the answers from Freebase IDs to entity names and paring them with English Wikipedia as an external knowledge source.
Using this format, WebQuestions is used as a benchmark by ODQA systems, e,g., R3\cite{Wang2018R3RR}, ORQA\cite{Lee2019LatentRF}, REALM\cite{Guu2020REALMRL}, DPR\cite{karpukhin-etal-2020-dense}, RAG\cite{Lewis2020RetrievalAugmentedGF}, GPT-3\cite{Brown2020LanguageMA}, T5\cite{roberts-etal-2020-much}, etc.

\subsection{WikiMovies}
WikiMovies\cite{Miller2016KeyValueMN} contains 96K movie-related question-answer pairs curated from OMDb and the MovieLens database.
All the questions in this dataset can be answered using a subset of Wikipedia pages.
WikiMovies was used as an ODQA benchmark by \cite{chen-etal-2017-reading} pairing it with English Wikipedia.
This format is also used in R3\cite{Wang2018R3RR}.

\subsection{OpenSQuAD}
OpenSQuAD is curated from the SQuAD 1.1\cite{Rajpurkar2016SQuAD1Q}, a popular Stanford MRC dataset annotated by crowd workers using Wikipedia passages.
It contains 87K examples for training and 10K for development. A large test set is sequestered by the authors for the leaderboard evaluation only.
Discarding context passages and using English Wikipedia, the question-answer pairs in the development set are used to evaluate ODQA systems.
This format is used in the following works: R3\cite{Wang2018R3RR}, DenSPI\cite{seo-etal-2019-real}, Multi-passage BERT\cite{wang-etal-2019-multi}, ORQA\cite{Lee2019LatentRF}, DPR\cite{karpukhin-etal-2020-dense}, Fusion-in-Decoder\cite{izacard-grave-2021-leveraging}, etc. 
However, \cite{karpukhin-etal-2020-dense} argues that SQuAD is not ideal for ODQA because “many questions lack context in the absence of the provided paragraphs”. Following the publication of \cite{karpukhin-etal-2020-dense} fewer works make use of OpenSQuAD.

\subsection{Qusar-T}
Qusar-T\cite{Dhingra2017QuasarDF} is a large-scale dataset designed for ODQA.
It contains 43K trivia question-answer pairs acquired from various internet sources.
Paired with English Wikipedia as the external source of knowledge, it is used as one of the benchmark datasets in recent ODQA works like R3\cite{Wang2018R3RR} and Multi-passage BERT\cite{wang-etal-2019-multi}.

\subsection{OpenTriviaQA}
TriviaQA was proposed by \cite{Joshi2017TriviaQAAL} as an MRC dataset containing over 650K question-answer-evidence triples whith 96K unique question-answer pairs collected from trivia websites and on average six evidence documents per question.
In ORQA\cite{Lee2019LatentRF}, paired with the 2018-12-20 dump of English Wikipedia as an external knowledge source, the unfiltered version of TriviaQA is used as an ODQA benchmark with the evidence documents removed.
This setting(denoted as OpenTriviaQA) is also used as a ODQA benchmark in the works of Multi-passage BERT\cite{wang-etal-2019-multi}, DPR\cite{karpukhin-etal-2020-dense} , RAG\cite{Lewis2020RetrievalAugmentedGF}, GPT-3\cite{Brown2020LanguageMA}, T5\cite{roberts-etal-2020-much}, Fusion-in-Decoder\cite{izacard-grave-2021-leveraging}, etc.

\subsection{OpenNaturalQuestions}
Natural Questions is Google’s MRC dataset. Its questions are gathered from user query logs of the Google search engine\cite{Kwiatkowski2019NaturalQA}. Long-form and short-form answers are annotated by crowd workers as a span of words based on the top-5 Wikipedia documents returned by the search engine. It contains 307K (question, Wikipedia page, long answer, short answer) quadruples in the training set and 8K in the development set. ORQA\cite{Lee2019LatentRF} make Natural Questions usable as an ODQA benchmark by only keeping the questions with short answers (with 5 tokens at most), discarding the evidence Wikipedia page, and instead using English Wikipedia as the only external knowledge source.
This format (denoted as OpenNaturalQuestions) is also followed by Multi-passage BERT\cite{wang-etal-2019-multi}, REALM\cite{Guu2020REALMRL}, DPR\cite{karpukhin-etal-2020-dense}, RAG\cite{Lewis2020RetrievalAugmentedGF}, GPT-3\cite{Brown2020LanguageMA}, T5\cite{roberts-etal-2020-much}, Fusion-in-Decoder\cite{izacard-grave-2021-leveraging}, etc.

\subsection{KILT benchmarks}
KILT is a benchmark for Knowledge Intensive Language Tasks proposed by Facebook AI Research\cite{Petroni2021KILTAB}.
It aims to break the boundaries between datasets for knowledge-intensive language tasks, including Slot Filling, ODQA, Dialogue, Fact Checking, and Entity Linking by mapping them to a unified snapshot of an external knowledge source (2019-08-01 dump of English Wikipedia).
Another important feature of KILT is that the format of the KILT benchmark is model-agnostic, which means it supports any system that takes textual input and produces textual output. This feature dramatically reduces the data preprocessing workload when testing across multiple datasets.
KILT also tries to provide as much provenance information as possible for all its instances through crowdsourcing. This helps people evaluate whether their system can provide correct evidence for their prediction.
Next, we will describe the four ODQA datasets included in KILT, which are KILT-TriviaQA, KILT-NaturalQuestions, KILT-HotpotQA, and KILT-ELI5.

\subsubsection{KILT-Natural Questions}
Spans of both long and short answers in the Natural Questions(NQ)\cite{Kwiatkowski2019NaturalQA} are used as provenance, and the evidence documents are discarded under the ODQA setting.
A new test set for KILT consists of the unpublished portion of NQ is acquired through collaboration with NQ's authors.

\subsubsection{KILT-HotpotQA}
The supporting sentences provided for each question-answer pair in the original HotpotQA\cite{Yang2018HotpotQAAD} are used as provenance. KILT uses a full-wiki setting where systems need to retrieve and reason over all of Wikipedia.

\subsubsection{KILT-TriviaQA}
KILT-TriviaQA only uses the question-answer pairs in the TriviaQA\cite{Joshi2017TriviaQAAL} that have an evidence context from Wikipedia. 
The span of the answer is used as provenance and the full version of the development set and the test set of TriviaQA are kept.

\subsubsection{KILT-EMLI5}
In the original EMLI5\cite{Fan2019ELI5LF}, a collection of web pages acquired from Common Crawl is provided for each question-answer pair as evidence documents.
In KILT, gold provenances are annotated for the development and test set of the original EMLI5 by asking turkers to select the evidence Wikipedia pages that actually contain the answer.

\section{Commonsense Question Answering}\label{sec:Commonsense QA}
Commonsense QA is a distinct task that cannot simply be classified as MRC or ODQA. Unlike MRC and ODQA it emphasises on those questions that are answerable by humans using commonsense reasoning even in the absence of the context information, external knowledge (e.g., SWAG, CODAH\cite{chen-etal-2019-codah}, CommonsenseQA\cite{Talmor2019CommonsenseQAAQ}), or context that explicitly contains the correct answer (e.g., COSMOS\cite{Huang2019CosmosQM}, ReCoRD\cite{Zhang2018ReCoRDBT}). Key features of each commonsense QA dataset are summarized in Table~\ref{tab:Commonsense}.

\begin{table*}[htbp]
    \centering
    \captionsetup{justification=centering}
    \caption{Commonsense Question Answering benchmarks.} \label{tab:Commonsense}%
    \setlength{\tabcolsep}{3pt} 
    \renewcommand{\arraystretch}{1.5} 
    \begin{tabular}{@{\extracolsep{3pt}}*1l@{} @{\extracolsep{3pt}}*6c@{} @{\extracolsep{3pt}}*1r@{}}
        \toprule
        
        &&& \multicolumn{3}{c}{\textbf{Source}} \\\cline{5-6}
        \textbf{Dataset} & \textbf{year} & \textbf{Context} & \textbf{\#Questions} & \textbf{Passages} & \textbf{QA pairs} & \textbf{Answer Type} & \textbf{Metrics}\\
        \hline
         &&&& Without Context Passage\\
        \midrule
       
        SWAG\cite{Zellers2018SWAGAL} & 2018 & Video/Movie caption & 113K & - & Synthetic Cloze & Multi-choice & Accuracy\\
        
        CODAH\cite{chen-etal-2019-codah} & 2019 & - & 2.8K & - & Crowdsourced & Multi-choice & Accuracy\\
        
        CommonsenseQA\cite{Talmor2019CommonsenseQAAQ} & 2019 & ConceptNet & 12K & - & Crowdsourced & Multi-choice & Accuracy\\

        \midrule
        &&&& With Context Passage\\
        \midrule
        
        ReCoRD\cite{Zhang2018ReCoRDBT} & 2018 & News & 120K & CNN/DM & Synthetic Cloze & Fill in entity & EM, F1\\
        &&&& Articles\\
        
        CosmosQA\cite{Huang2019CosmosQM} & 2019 & Blog & 33K & Spinn3r Dataset & Crowdsourced & Multi-choice & Accuracy\\
        &&& Articles\\
        \midrule
    \end{tabular}
\end{table*}%

\subsection{SWAG}
\begin{figure}
    \centering
    \includegraphics[width=\linewidth]{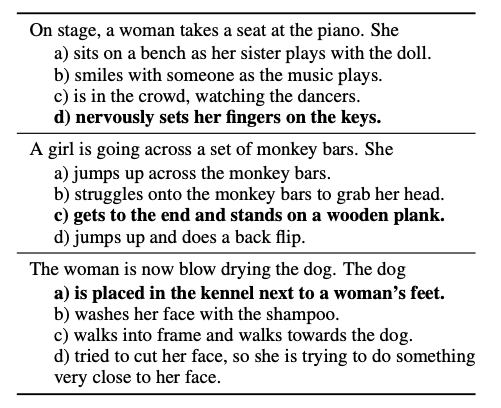}
    \caption{Examples from the SWAG dataset\cite{Zellers2018SWAGAL}.}
    \label{fig:SWAG}
\end{figure}

SWAG\cite{Zellers2018SWAGAL} is a large-scale cloze-style dataset designed for commonsense QA and proposed by researchers from the University of Washington. The task is formulated as, given the first sentence and the first noun of the second sentence, a system needs to select the correct ending of the second sentence from a collection of artificially generated endings. The artificial endings are generated by an RNN-based language model (LM) and then selected by an adversarial model to make sure those artificial endings in the pool cannot be easily identified. Finally, human annotators are hired to check those ending to ensure agreement. The corpus contains 113K multiple choice questions derived from consecutive video captions from two sources, the ActivityNet captions\cite{Krishna2017DenseCaptioningEI}\cite{Heilbron2015ActivityNetAL} and the Large Scale Movie Description Challenge\cite{Rohrbach2016MovieD}. Figure~\ref{fig:SWAG}, borrowed from \cite{Zellers2018SWAGAL}, gives examples of SWAG. As it shows, no context information is provided for each cloze-style question forcing systems to learn the underlying commonsense logic of humans solely through QA pairs. System performance is evaluated by accuracy.

\subsection{CODAH}
CODAH\cite{chen-etal-2019-codah} is a more challenging evaluation dataset extended from SWAG, proposed by researchers from Northwestern University. It contains 2.8K multi-choice sentence completion questions. Annotators are educated and rewarded to submit questions that adversarially target the weaknesses of a pre-trained model.

\subsection{CommonsenseQA}
\begin{figure}
    \centering
    \captionsetup{justification=raggedright,singlelinecheck=false}
    \includegraphics[width=\linewidth]{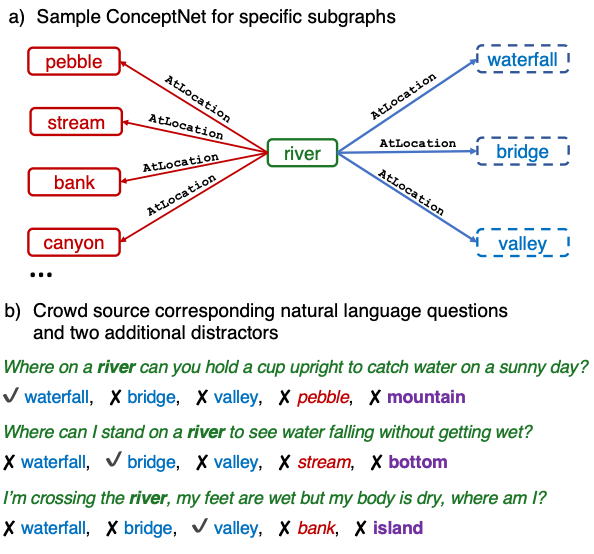}
    \caption{An illustration of a subgraph from ConceptNet and a few examples created based on it in the CommonsenseQA dataset\cite{Talmor2019CommonsenseQAAQ}.}
    \label{fig:CommonsenseQA}
\end{figure}

CommonsenseQA\cite{Talmor2019CommonsenseQAAQ} is a commonsense QA dataset proposed by researchers from the AI2 for question answering with prior knowledge. Crowd workers created 12K multi-choice questions from ConceptNet\cite{Speer2017ConceptNet5A} to capture common sense beyond associations. The questions in this dataset are supposed to be answerable using only commonsense knowledge. Figure~\ref{fig:CommonsenseQA}, borrowed from \cite{Talmor2019CommonsenseQAAQ}, gives examples from the CommonsenseQA dataset and shows how they are created based on a ConceptNet subgraph.

\subsection{ReCoRD}
\begin{figure}
    \centering
    \captionsetup{justification=raggedright,singlelinecheck=false}
    \includegraphics[width=\linewidth]{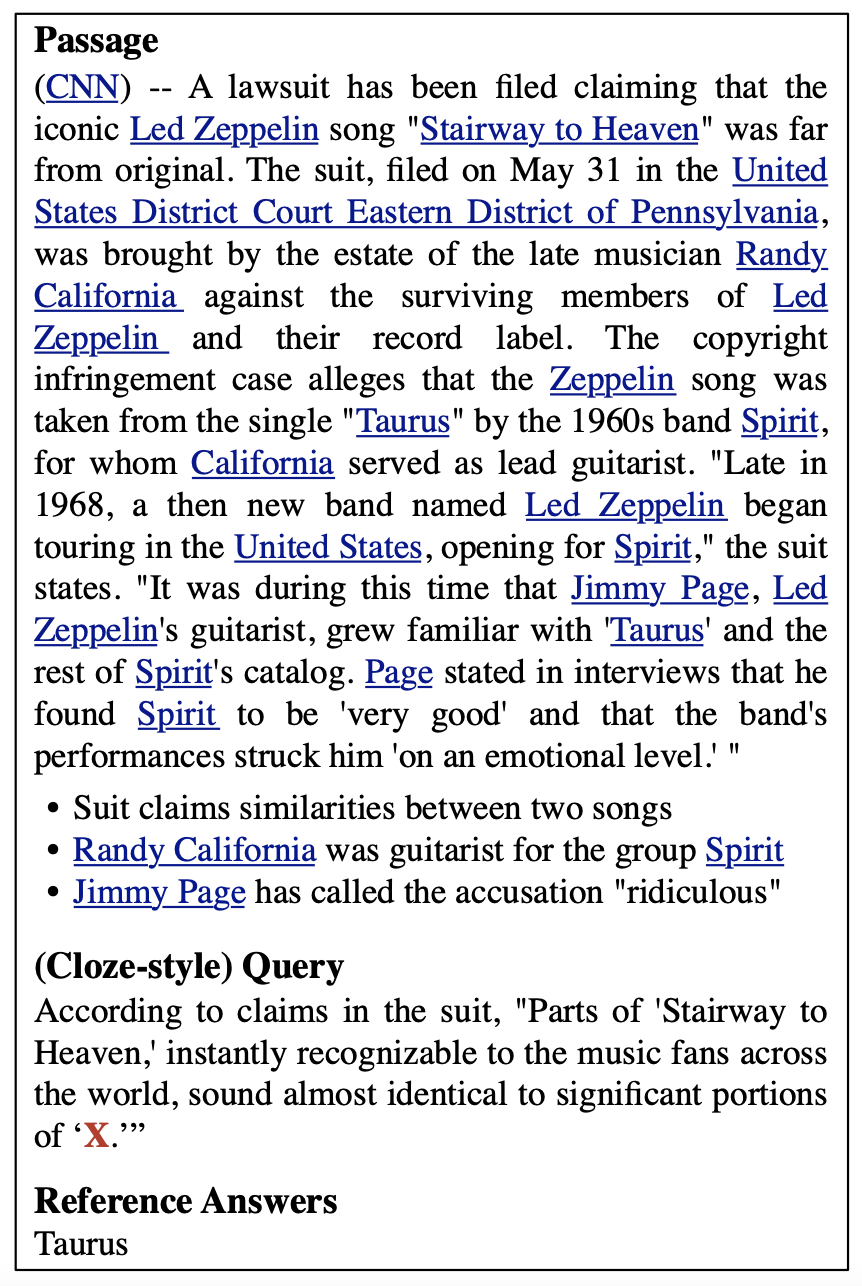}
    \caption{An example from the ReCoRD dataset\cite{Zhang2018ReCoRDBT}}
    \label{fig:ReCoRD}
\end{figure}

ReCoRD\cite{Zhang2018ReCoRDBT} is a large-scale Commonsense QA dataset proposed by researchers from Johns Hopkins University and Microsoft Research. It is presented in the form of a cloze-style MRC dataset that contains over 120K instances. Its creators claim that most of the questions require deep commonsense reasoning across the context passages to answer. Figure~\ref{fig:ReCoRD}, borrowed from \cite{Zhang2018ReCoRDBT}, illustrates an example of ReCoRD, where the 'X' in the query indicates a missing named entity that a system needs to infer from the context passage. The context passages are collected from CNN and Daily mail news articles. Specifically, the first few paragraphs of a news article together with the editor-provided highlights are sampled as the context passages because they usually summarize the article. The questions are constructed from sentences in the rest of the article by replacing a named entity with a placeholder, and usng the named entity as the correct answer. Five carefully designed criteria are applied when selecting the sentences to contract questions. See\cite{Zhang2018ReCoRDBT} for more details. EM and F1 measures are used for evaluation.

\subsection{CosmosQA}
\begin{figure}
    \centering
    \captionsetup{justification=raggedright,singlelinecheck=false}
    \includegraphics[width=\linewidth]{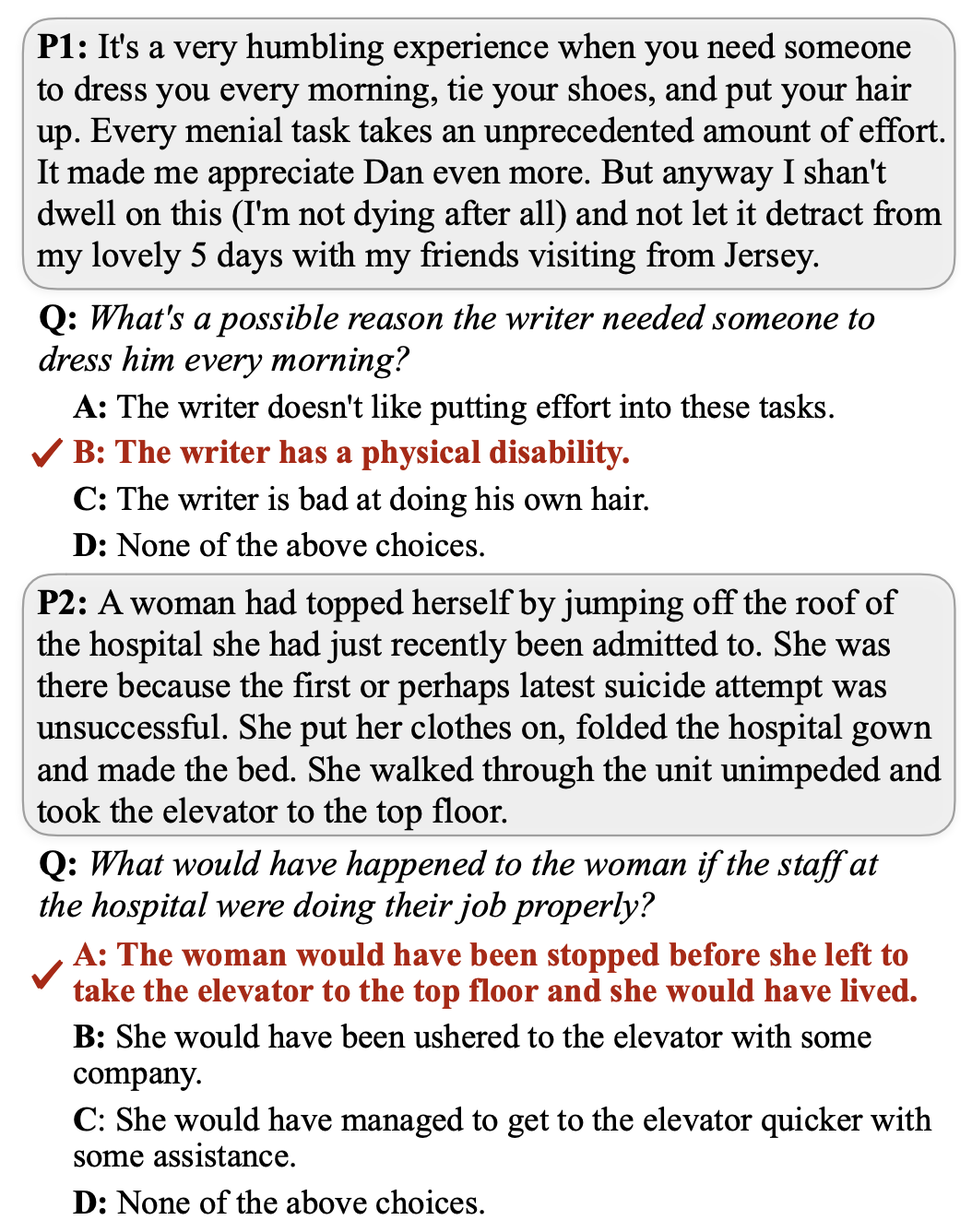}
    \caption{An example from the CosmosQA dataset\cite{Huang2019CosmosQM}.}
    \label{fig:CosmosQA}
\end{figure}

CosmosQA\cite{Huang2019CosmosQM} is a large-scale Commonsense QA dataset proposed by researchers from the UIUC and AI2. It can be also interpreted as a multi-choice style MRC dataset that includes 33K questions that require commonsense reasoning over the associated context passages. Figure~\ref{fig:CosmosQA}, borrowed from \cite{Huang2019CosmosQM}, gives two examples from the CosmosQA dataset. In CosmosQA, most of the correct answers are not explicitly presented in the context passages but can be inferred based on the content of the context. Such a configuration challenges textual QA systems MRC ability and their commonsense reasoning ability. The context passages are collected from a corpus of personal narratives from the Spinn3r Blog Dataset\cite{Burton2009TheI2}. Based on each context passage, up to two multi-choice questions are created by crowd workers asked to craft questions from four categories: causes of events, effects of events, facts about entities, and counterfactuals. The task is evaluated by accuracy.

\section{Metrics}\label{sec:Metrics}
\begin{figure}
    \centering
    \includegraphics[width=\linewidth]{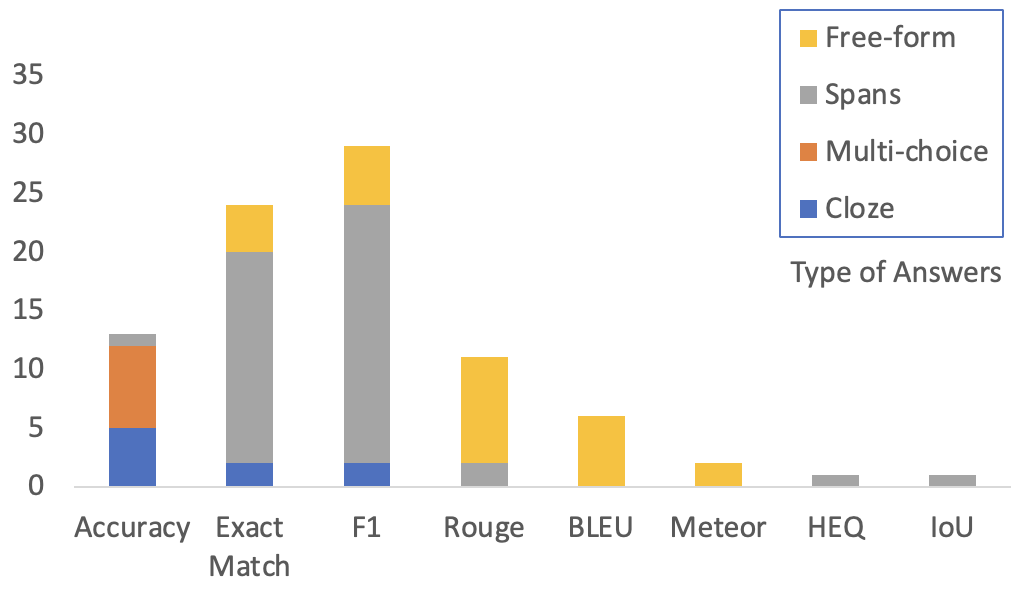}
    \caption{Distribution of metrics by answer types.}
    \label{fig:Metrics}
\end{figure}

In this section, we describe the metrics for the Textual QA tasks mentioned above. Figure~\ref{fig:Metrics} illustrates a metric distribution of datasets covered in this survey by answer types. Note that, for those ODQA benchmark datasets, we count CuratedTREC, WebQuestions, Qusar-T, and KILT-EMLI5 as free-form style datasets because their answers were not constructed using the words in an external knowledge source (i.e., Wikipedia), and the remaining ODQA benchmark datasets are counted as span extraction style datasets. From Figure~\ref{fig:Metrics} we can see that the preferred metric varies with the dataset type. Specifically, cloze-style and multi-choice style datasets commonly use accuracy for evaluation; Span extraction style datasets prefer to used EM and F1 as evaluation metrics. For the free-form style datasets, Rouge and BLEU are the most popular choices.

\subsection{Accuracy}
Accuracy represents the percentage of questions that a system answered correctly. Let \textbf{M} denote the number of questions a system answered correctly and \textbf{N} denote the total number of questions in the evaluation dataset. Accuracy is defined as:

\begin{equation} \label{eq:accuracy}
\rm Accuracy = \frac{M}{N}
\end{equation}

\subsection{Exact Match}
Exact Match (EM) measures the percentage of predictions that match any reference answers exactly at the token level (in some datasets, each question may have multiple reference answers). In other words, a prediction is counted as correct only when it exactly matches any one of the reference answers to the given question. Let \textbf{M} denote the number of correct predictions, \textbf{N} denote the total number of questions in the evaluation dataset. EM is defined in Eq.~\ref{eq:EM}:

\begin{equation} \label{eq:EM}
\rm EM = \frac{M}{N}
\end{equation}

It is worth noting that for span extraction tasks, EM is the same as Accuracy.

\subsection{F1}
The F1 score is the harmonic mean of the precision and the recall. In textual QA, the precision measures the ratio of the number of tokens in a prediction that overlap with the correct answer to the total number of tokens in the prediction. The recall measures the ratio of the number of tokens in a correct answer that have been covered by a prediction to the total number of tokens in the correct answer. Equations of token-level precision and recall for a single prediction are shown in Eq.~\ref{eq:Precision} and Eq.~\ref{eq:Recall} respectively.

\begin{equation} \label{eq:Precision}
\rm Precision = \frac{TP}{TP + FP}
\end{equation}

\begin{equation} \label{eq:Recall}
\rm Recall = \frac{TP}{TP + FN}
\end{equation}

where TP, FP, FN denote the token-level true positive, false positive, and false negative respectively. The equation of the F1 score for a single prediction is given in Eq.~\ref{eq:F1}:

\begin{equation} \label{eq:F1}
\rm F1 = 2 * \frac{Precision * Recall}{Precision + Recall}
\end{equation}

When a question has multiple reference answers, the highest F1 score is taken. The average F1 score of all predictions is the final F1 score of the system.

\subsection{ROUGE}
ROUGE\cite{Lin2004ROUGEAP} is the abbreviation for Recall-Oriented Understudy for Gisting Evaluation. It consists of a set of metrics that were originally proposed to evaluate automatic text summarization. Because it works by comparing the automatically generated text with the human-generated reference text, it is now widely used by textual QA benchmarks especially by those with free-form reference answers.

The most popular ROUGE metrics in textual QA are ROUGE-N and ROUGE-L, which represent a comparison of texts at different granularities. Specifically, ROUGE-N measures the ratio of the number of overlapped n-grams between the generated text and the reference text to the total n-grams in the reference text and is defined in Eq.~\ref{eq:ROUGE-N}. Similarly, ROUGE-L measures the longest matching sequence of words using the longest common subsequence (LCS) which can automatically include the longest in-sequence n-grams.

\begin{equation} \label{eq:ROUGE-N}
\begin{aligned}
\rm ROU {} & \rm GE_N = \\
& \rm \frac{\sum_{snt\in{Ref}}\sum_{gram_n\in{snt}}Count_{match}(gram_n)}{\sum_{snt\in{Ref}}\sum_{gram_n\in{snt}}Count(gram_n)}
\end{aligned}
\end{equation}

\subsection{BLEU}
BLEU\cite{Papineni2002BleuAM} stands for Bilingual Evaluation Understudy, which was originally proposed to evaluate machine translations. Similar to ROUGE, BLEU compares the lexical features between two sequences to words, and it is also widely used for generative textual QA evaluations. Specifically, BLEU is defined as follows:

\begin{equation}  \label{eq:BLEU-Pn}
\rm p_n = \frac{\sum_{sent\in{C}}\sum_{gram_n\in{sent}}min(m_{cand}^{gram_n}, m_{ref}^{gram_n})}{w_t^{gram_n} = \sum_{snt'\in{C}}\sum_{gram_n'\in{snt'}m_{cand}^{gram_n'}}}
\end{equation}

\begin{equation} \label{eq:BLEU}
\begin{aligned}
\rm {} & \rm BLEU = \\
& \rm \underbrace{\min(1, \exp(1 - \rm \frac{reference\_length}{output\_length}))}_{brevity\  penalty}\underbrace{(\prod_{i=1}^4 \rm p_n)^{1/4}}_{n-gram\ overlap}
\end{aligned}
\end{equation}

where $p_n$ is called n-gram precision in which $m_{cand}^{gram_n}$ is the number of the n-gram in the prediction(candidate) matching the reference answer;
$C$ is the candidate answer set;
$m_{ref}^{gram_n}$ is the number of the n-gram in the reference answer; $w_t^{gram_n}$ is the total number of the n-grams in the prediction.

\subsection{Meteor}
Meteor\cite{banerjee-lavie-2005-meteor} is another automatic metric that was originally proposed for machine translation. It claims to have a better correlation with human judgment. Similar to BLEU, Meteor also consists of two parts: a weighted F score and a penalty factor, which is given by the following equations: 

\begin{equation}  \label{eq:Meteor}
\rm Meteor = F * (1 - Penalty)
\end{equation}

with

\begin{equation}  \label{eq:Meteor-F}
\rm F = \frac{Precision * Recall}{\alpha * Precision + (1 - \alpha) * Recall}
\end{equation}

\begin{equation}  \label{eq:Meteor-Penalty}
\rm Penalty = \gamma * (\frac{ch}{m})^{\beta}
\end{equation}

where $ch$ is the total number of matching chunks and m is the total number of matched uniforms between the prediction and the reference. The number of chunks decreases when the matches are contiguous which will reduce the penalty. The precision and recall are given as $m/c$ and $m/r$, in which $c$ and $r$ are the lengths of the candidate and reference respectively. Finally, the parameters $\alpha$, $\beta$, and $\gamma$ are tuned to maximize the correlation with human judgments.

\subsection{HEQ}
HEQ stands for Human Equivalence Score, which was proposed along with the conversational QA benchmark dataset, QuAC\cite{Choi2018QuACQA}. HEQ creators claim that the F1 score can be misleading for questions with multiple answers. HEQ measures the percentage of examples for which the F1 score of predictions exceeds or matches the human F1 score. HEQ has two variants: HEQ-Q (the percentage of questions for which this is true) and HEQ-D (the percentage of dialogues for which this is true for every question in the dialog).

\subsection{IoU}
IoU stands for Intersection over Union. It was used with the long-form QA benchmark dataset, NLQuAD\cite{Soleimani2021NLQuADAN}. NLQuAD creators claim that F1 score and ROUGE-N have limited use for evaluation of long answers. IoU measures position-sensitive overlap between two spans which is defined as follows:

\begin{equation}  \label{eq:IoU}
\rm IoU = \frac{|p \bigcap t|}{|p \bigcup t|}
\end{equation}

where $p$ and $t$ are the predicted and target contiguous intervals over the context document, containing the positions of the tokens.
\section{Current Trends and Future Directions}\label{sec:Trends&Directions}
In this section, we summarize the current trends in constructing textual QA tasks suggestions for future textual QA benchmark construction.

\subsection{Current Trends}
For the question collection methods, we see a trend of collecting real-life user questions from either user logs of search engines or QA websites like Reddit, Stack overflow, etc., instead of constructing cloze-style queries automatically or crowdsourcing questions based on a context passage/document. 

For answer collection methods, the free-form answers and span-style answers are the two most popular answer type choices. The free-form answers are usually directly derived from QA websites, while span-style answers are mostly collected through crowd workers, which is more expensive. In addition, we see a tendency towards annotating long answers, thus providing more supportive information to questions. Moreover, in some of the latest datasets, supportive sentences in the context documents are annotated for each answer in order to help increase the interpretability of the system by requiring it not only to return the answer but also these evidence sentences.

For the selection of context of documents or external knowledge sources, we find that Wikipedia is the most popular choice, especially for ODQA tasks. A few MRC datasets choose other domains like news, movies, HealthCare, Twitter, etc.

\subsection{Future Directions}
We believe providing evidence sentences for each answer is a good direction for future textual QA datasets. Besides evidence sentences, we hope to see the use of other forms of information that can support answer selection, e.g., pictures and hyperlinks in the document. This can be of benefit to future research beyond single-model textual QA.

We recommend that instead of continuing to propose new datasets based on Wikipedia articles, researchers should explore other information domains, such as healthcare or social media, each of which has different data quality requirements. For example, in the healthcare domain, people expect high precision, while in the social media domain, people may be more tolerant of misinformation. Different challenges may be constructed based on these features.

For metrics, as we discussed in section~\ref{sec:Metrics}, different textual QA tasks have various flavors of evaluation measures. But do we really need all of them? We think evaluating and unifying these metrics for textual QA would be a meaningful future work. This not only can reduce the burden when people need to evaluate their models over multiple benchmark datasets but also help people have a better understanding of key features when comparing two answers in natural language. This in turn can help coach the development of textual QA systems. We also believe that more metrics should be evaluated for textual QA tasks. We make this suggestion because we noticed that recently there are many new metrics proposed for text generation tasks, such as the BERT-score\cite{Zhang2020BERTScoreET} and the BLEURT\cite{Sellam2020BLEURTLR}. Both claim to have better alignment with human evaluation than lexical feature-based metrics like ROUGE, BLEU, and Meteor that are widely used in textual QA tasks currently.

Finally, we find that most of the benchmark datasets only have one reference answer for each question, and all of them only require a system to return one answer to each question. In fact, in realistic scenarios, many questions may have multiple answers based on different supportive knowledge. For example, in social media, people may be interested in an event that may not have a certain answer but only a few popular rumors available. Accordingly, new metrics for multiple-answer QA need to be developed. 

\section{Conclusion}\label{sec:Conclusion}

In this survey, we reviewed 47 textual QA benchmark datasets and their corresponding evaluation metrics. A novel taxonomy of textual QA tasks is provided from an application scenario point of view. A detailed description is provided for each dataset, which covers the task definition, contraction method, statistics, and evaluation measures. Key features and detailed statistics among the benchmark datasets are summarized and compared in the form of tables. Detailed description and distribution analysis of evaluation metrics are provided. Finally, we summarized the trends of the recent textual QA benchmark contraction methods and give our opinions on the future directions of the textual QA benchmark research. We hope this work serves as a good introduction to textual QA tasks.




\ifCLASSOPTIONcompsoc
  \section*{Acknowledgments}
\else
  \section*{Acknowledgment}
\fi

We thank Ira Harmon from the University of Florida for his feedback to this paper.

\ifCLASSOPTIONcaptionsoff
  \newpage
\fi

\bibliography{IEEEtran.bib}{}

\begin{thebibliography}{10}
\providecommand{\url}[1]{#1}
\csname url@samestyle\endcsname
\providecommand{\newblock}{\relax}
\providecommand{\bibinfo}[2]{#2}
\providecommand{\BIBentrySTDinterwordspacing}{\spaceskip=0pt\relax}
\providecommand{\BIBentryALTinterwordstretchfactor}{4}
\providecommand{\BIBentryALTinterwordspacing}{\spaceskip=\fontdimen2\font plus
\BIBentryALTinterwordstretchfactor\fontdimen3\font minus
  \fontdimen4\font\relax}
\providecommand{\BIBforeignlanguage}[2]{{%
\expandafter\ifx\csname l@#1\endcsname\relax
\typeout{** WARNING: IEEEtran.bst: No hyphenation pattern has been}%
\typeout{** loaded for the language `#1'. Using the pattern for}%
\typeout{** the default language instead.}%
\else
\language=\csname l@#1\endcsname
\fi
#2}}
\providecommand{\BIBdecl}{\relax}
\BIBdecl

\bibitem{Green1961BaseballAA}
B.~Green, A.~K. Wolf, C.~Chomsky, and K.~Laughery, ``Baseball: an automatic
  question-answerer,'' in \emph{IRE-AIEE-ACM '61 (Western)}, 1961.

\bibitem{Voorhees2000BuildingAQ}
E.~Voorhees and D.~M. Tice, ``Building a question answering test collection,''
  in \emph{SIGIR '00}, 2000.

\bibitem{Bollacker2008FreebaseAC}
K.~Bollacker, C.~Evans, P.~K. Paritosh, T.~Sturge, and J.~Taylor, ``Freebase: a
  collaboratively created graph database for structuring human knowledge,'' in
  \emph{SIGMOD Conference}, 2008.

\bibitem{Suchanek2007YagoAC}
F.~M. Suchanek, G.~Kasneci, and G.~Weikum, ``Yago: a core of semantic
  knowledge,'' in \emph{WWW '07}, 2007.

\bibitem{devlin-etal-2019-bert}
\BIBentryALTinterwordspacing
J.~Devlin, M.-W. Chang, K.~Lee, and K.~Toutanova, ``{BERT}: Pre-training of
  deep bidirectional transformers for language understanding,'' in
  \emph{Proceedings of the 2019 Conference of the North {A}merican Chapter of
  the Association for Computational Linguistics: Human Language Technologies,
  Volume 1 (Long and Short Papers)}.\hskip 1em plus 0.5em minus 0.4em\relax
  Minneapolis, Minnesota: Association for Computational Linguistics, Jun. 2019,
  pp. 4171--4186. [Online]. Available: \url{https://aclanthology.org/N19-1423}
\BIBentrySTDinterwordspacing

\bibitem{Brown2020LanguageMA}
T.~B. Brown, B.~Mann, N.~Ryder, M.~Subbiah, J.~Kaplan, P.~Dhariwal,
  A.~Neelakantan, P.~Shyam, G.~Sastry, A.~Askell, S.~Agarwal, A.~Herbert-Voss,
  G.~Krueger, T.~Henighan, R.~Child, A.~Ramesh, D.~M. Ziegler, J.~Wu,
  C.~Winter, C.~Hesse, M.~Chen, E.~Sigler, M.~Litwin, S.~Gray, B.~Chess,
  J.~Clark, C.~Berner, S.~McCandlish, A.~Radford, I.~Sutskever, and D.~Amodei,
  ``Language models are few-shot learners,'' \emph{ArXiv}, vol. abs/2005.14165,
  2020.

\bibitem{lewis-etal-2020-bart}
\BIBentryALTinterwordspacing
M.~Lewis, Y.~Liu, N.~Goyal, M.~Ghazvininejad, A.~Mohamed, O.~Levy, V.~Stoyanov,
  and L.~Zettlemoyer, ``{BART}: Denoising sequence-to-sequence pre-training for
  natural language generation, translation, and comprehension,'' in
  \emph{Proceedings of the 58th Annual Meeting of the Association for
  Computational Linguistics}.\hskip 1em plus 0.5em minus 0.4em\relax Online:
  Association for Computational Linguistics, Jul. 2020, pp. 7871--7880.
  [Online]. Available: \url{https://aclanthology.org/2020.acl-main.703}
\BIBentrySTDinterwordspacing

\bibitem{Raffel2020ExploringTL}
C.~Raffel, N.~M. Shazeer, A.~Roberts, K.~Lee, S.~Narang, M.~Matena, Y.~Zhou,
  W.~Li, and P.~J. Liu, ``Exploring the limits of transfer learning with a
  unified text-to-text transformer,'' \emph{ArXiv}, vol. abs/1910.10683, 2020.

\bibitem{Dzendzik2021EnglishMR}
D.~Dzendzik, C.~Vogel, and J.~Foster, ``English machine reading comprehension
  datasets: A survey,'' \emph{ArXiv}, vol. abs/2101.10421, 2021.

\bibitem{Saeidi2018InterpretationON}
M.~Saeidi, M.~Bartolo, P.~Lewis, S.~Singh, T.~Rockt{\"a}schel, M.~Sheldon,
  G.~Bouchard, and S.~Riedel, ``Interpretation of natural language rules in
  conversational machine reading,'' in \emph{EMNLP}, 2018.

\bibitem{Choi2018QuACQA}
E.~Choi, H.~He, M.~Iyyer, M.~Yatskar, W.~tau Yih, Y.~Choi, P.~Liang, and
  L.~Zettlemoyer, ``Quac: Question answering in context,'' in \emph{EMNLP},
  2018.

\bibitem{Reddy2019CoQAAC}
S.~Reddy, D.~Chen, and C.~D. Manning, ``Coqa: A conversational question
  answering challenge,'' \emph{Transactions of the Association for
  Computational Linguistics}, vol.~7, pp. 249--266, 2019.

\bibitem{Yang2018HotpotQAAD}
Z.~Yang, P.~Qi, S.~Zhang, Y.~Bengio, W.~W. Cohen, R.~Salakhutdinov, and C.~D.
  Manning, ``Hotpotqa: A dataset for diverse, explainable multi-hop question
  answering,'' in \emph{EMNLP}, 2018.

\bibitem{Ho2020ConstructingAM}
X.~Ho, A.~Nguyen, S.~Sugawara, and A.~Aizawa, ``Constructing a multi-hop qa
  dataset for comprehensive evaluation of reasoning steps,'' in \emph{COLING},
  2020.

\bibitem{Chen2020HybridQAAD}
W.~Chen, H.~Zha, Z.~Chen, W.~Xiong, H.~Wang, and W.~Wang, ``Hybridqa: A dataset
  of multi-hop question answering over tabular and textual data,'' in
  \emph{FINDINGS}, 2020.

\bibitem{Fan2019ELI5LF}
A.~Fan, Y.~Jernite, E.~Perez, D.~Grangier, J.~Weston, and M.~Auli, ``Eli5: Long
  form question answering,'' in \emph{ACL}, 2019.

\bibitem{zhu2020question}
M.~Zhu, A.~Ahuja, D.-C. Juan, W.~Wei, and C.~K. Reddy, ``Question answering
  with long multiple-span answers,'' in \emph{Proceedings of the 2020
  Conference on Empirical Methods in Natural Language Processing: Findings},
  2020, pp. 3840--3849.

\bibitem{Soleimani2021NLQuADAN}
A.~Soleimani, C.~Monz, and M.~Worring, ``Nlquad: A non-factoid long question
  answering data set,'' in \emph{EACL}, 2021.

\bibitem{liu-etal-2019-xqa}
\BIBentryALTinterwordspacing
J.~Liu, Y.~Lin, Z.~Liu, and M.~Sun, ``{XQA}: A cross-lingual open-domain
  question answering dataset,'' in \emph{Proceedings of the 57th Annual Meeting
  of the Association for Computational Linguistics}.\hskip 1em plus 0.5em minus
  0.4em\relax Florence, Italy: Association for Computational Linguistics, Jul.
  2019, pp. 2358--2368. [Online]. Available:
  \url{https://aclanthology.org/P19-1227}
\BIBentrySTDinterwordspacing

\bibitem{artetxe-etal-2020-cross}
\BIBentryALTinterwordspacing
M.~Artetxe, S.~Ruder, and D.~Yogatama, ``On the cross-lingual transferability
  of monolingual representations,'' in \emph{Proceedings of the 58th Annual
  Meeting of the Association for Computational Linguistics}.\hskip 1em plus
  0.5em minus 0.4em\relax Online: Association for Computational Linguistics,
  Jul. 2020, pp. 4623--4637. [Online]. Available:
  \url{https://aclanthology.org/2020.acl-main.421}
\BIBentrySTDinterwordspacing

\bibitem{lewis-etal-2020-mlqa}
\BIBentryALTinterwordspacing
P.~Lewis, B.~Oguz, R.~Rinott, S.~Riedel, and H.~Schwenk, ``{MLQA}: Evaluating
  cross-lingual extractive question answering,'' in \emph{Proceedings of the
  58th Annual Meeting of the Association for Computational Linguistics}.\hskip
  1em plus 0.5em minus 0.4em\relax Online: Association for Computational
  Linguistics, Jul. 2020, pp. 7315--7330. [Online]. Available:
  \url{https://aclanthology.org/2020.acl-main.653}
\BIBentrySTDinterwordspacing

\bibitem{chen-etal-2017-reading}
\BIBentryALTinterwordspacing
D.~Chen, A.~Fisch, J.~Weston, and A.~Bordes, ``Reading {W}ikipedia to answer
  open-domain questions,'' in \emph{Proceedings of the 55th Annual Meeting of
  the Association for Computational Linguistics (Volume 1: Long Papers)}.\hskip
  1em plus 0.5em minus 0.4em\relax Vancouver, Canada: Association for
  Computational Linguistics, Jul. 2017, pp. 1870--1879. [Online]. Available:
  \url{https://aclanthology.org/P17-1171}
\BIBentrySTDinterwordspacing

\bibitem{Lee2019LatentRF}
K.~Lee, M.-W. Chang, and K.~Toutanova, ``Latent retrieval for weakly supervised
  open domain question answering,'' in \emph{ACL}, 2019.

\bibitem{Wang2018R3RR}
S.~Wang, M.~Yu, X.~Guo, Z.~Wang, T.~Klinger, W.~Zhang, S.~Chang, G.~Tesauro,
  B.~Zhou, and J.~Jiang, ``R3: Reinforced ranker-reader for open-domain
  question answering,'' in \emph{AAAI}, 2018.

\bibitem{Petroni2021KILTAB}
F.~Petroni, A.~Piktus, A.~Fan, P.~Lewis, M.~Yazdani, N.~D. Cao, J.~Thorne,
  Y.~Jernite, V.~Plachouras, T.~Rocktaschel, and S.~Riedel, ``Kilt: a benchmark
  for knowledge intensive language tasks,'' in \emph{NAACL}, 2021.

\bibitem{Zellers2018SWAGAL}
R.~Zellers, Y.~Bisk, R.~Schwartz, and Y.~Choi, ``Swag: A large-scale
  adversarial dataset for grounded commonsense inference,'' in \emph{EMNLP},
  2018.

\bibitem{chen-etal-2019-codah}
\BIBentryALTinterwordspacing
M.~Chen, M.~D{'}Arcy, A.~Liu, J.~Fernandez, and D.~Downey, ``{CODAH}: An
  adversarially-authored question answering dataset for common sense,'' in
  \emph{Proceedings of the 3rd Workshop on Evaluating Vector Space
  Representations for {NLP}}.\hskip 1em plus 0.5em minus 0.4em\relax
  Minneapolis, USA: Association for Computational Linguistics, Jun. 2019, pp.
  63--69. [Online]. Available: \url{https://aclanthology.org/W19-2008}
\BIBentrySTDinterwordspacing

\bibitem{Talmor2019CommonsenseQAAQ}
A.~Talmor, J.~Herzig, N.~Lourie, and J.~Berant, ``Commonsenseqa: A question
  answering challenge targeting commonsense knowledge,'' in \emph{NAACL}, 2019.

\bibitem{Huang2019CosmosQM}
L.~Huang, R.~L. Bras, C.~Bhagavatula, and Y.~Choi, ``Cosmos qa: Machine reading
  comprehension with contextual commonsense reasoning,'' in
  \emph{EMNLP/IJCNLP}, 2019.

\bibitem{Zhang2018ReCoRDBT}
S.~Zhang, X.~Liu, J.~Liu, J.~Gao, K.~Duh, and B.~V. Durme, ``Record: Bridging
  the gap between human and machine commonsense reading comprehension,''
  \emph{ArXiv}, vol. abs/1810.12885, 2018.

\bibitem{Cambazoglu2020ARO}
B.~B. Cambazoglu, M.~Sanderson, F.~Scholer, and B.~Croft, ``A review of public
  datasets in question answering research,'' \emph{ACM SIGIR Forum}, vol.~54,
  pp. 1 -- 23, 2020.

\bibitem{Zeng2020ASO}
C.~Zeng, S.~Li, Q.~Li, J.~Hu, and J.~Hu, ``A survey on machine reading
  comprehension: Tasks, evaluation metrics, and benchmark datasets,''
  \emph{ArXiv}, vol. abs/2006.11880, 2020.

\bibitem{Zhu2021RetrievingAR}
F.~Zhu, W.~Lei, C.~Wang, J.~Zheng, S.~Poria, and T.-S. Chua, ``Retrieving and
  reading: A comprehensive survey on open-domain question answering,''
  \emph{ArXiv}, vol. abs/2101.00774, 2021.

\bibitem{Zaib2021ConversationalQA}
M.~Zaib, W.~E. Zhang, Q.~Z. Sheng, A.~Mahmood, and Y.~Zhang, ``Conversational
  question answering: A survey,'' \emph{ArXiv}, vol. abs/2106.00874, 2021.

\bibitem{Yang2019XLNetGA}
Z.~Yang, Z.~Dai, Y.~Yang, J.~Carbonell, R.~Salakhutdinov, and Q.~V. Le,
  ``Xlnet: Generalized autoregressive pretraining for language understanding,''
  in \emph{NeurIPS}, 2019.

\bibitem{hermann2015teaching}
K.~M. Hermann, T.~Kocisky, E.~Grefenstette, L.~Espeholt, W.~Kay, M.~Suleyman,
  and P.~Blunsom, ``Teaching machines to read and comprehend,'' \emph{Advances
  in neural information processing systems}, vol.~28, pp. 1693--1701, 2015.

\bibitem{cui-etal-2016-consensus}
\BIBentryALTinterwordspacing
Y.~Cui, T.~Liu, Z.~Chen, S.~Wang, and G.~Hu, ``Consensus attention-based neural
  networks for {C}hinese reading comprehension,'' in \emph{Proceedings of
  {COLING} 2016, the 26th International Conference on Computational
  Linguistics: Technical Papers}.\hskip 1em plus 0.5em minus 0.4em\relax Osaka,
  Japan: The COLING 2016 Organizing Committee, Dec. 2016, pp. 1777--1786.
  [Online]. Available: \url{https://aclanthology.org/C16-1167}
\BIBentrySTDinterwordspacing

\bibitem{Hill2016TheGP}
F.~Hill, A.~Bordes, S.~Chopra, and J.~Weston, ``The goldilocks principle:
  Reading children's books with explicit memory representations,'' \emph{CoRR},
  vol. abs/1511.02301, 2016.

\bibitem{Bajgar2016EmbracingDA}
O.~Bajgar, R.~Kadlec, and J.~Kleindienst, ``Embracing data abundance: Booktest
  dataset for reading comprehension,'' \emph{ArXiv}, vol. abs/1610.00956, 2016.

\bibitem{richardson-etal-2013-mctest}
\BIBentryALTinterwordspacing
M.~Richardson, C.~J. Burges, and E.~Renshaw, ``{MCT}est: A challenge dataset
  for the open-domain machine comprehension of text,'' in \emph{Proceedings of
  the 2013 Conference on Empirical Methods in Natural Language
  Processing}.\hskip 1em plus 0.5em minus 0.4em\relax Seattle, Washington, USA:
  Association for Computational Linguistics, Oct. 2013, pp. 193--203. [Online].
  Available: \url{https://aclanthology.org/D13-1020}
\BIBentrySTDinterwordspacing

\bibitem{Tapaswi2016MovieQAUS}
M.~Tapaswi, Y.~Zhu, R.~Stiefelhagen, A.~Torralba, R.~Urtasun, and S.~Fidler,
  ``Movieqa: Understanding stories in movies through question-answering,''
  \emph{2016 IEEE Conference on Computer Vision and Pattern Recognition
  (CVPR)}, pp. 4631--4640, 2016.

\bibitem{Lai2017RACELR}
G.~Lai, Q.~Xie, H.~Liu, Y.~Yang, and E.~Hovy, ``Race: Large-scale reading
  comprehension dataset from examinations,'' in \emph{EMNLP}, 2017.

\bibitem{Rajpurkar2016SQuAD1Q}
P.~Rajpurkar, J.~Zhang, K.~Lopyrev, and P.~Liang, ``Squad: 100,000+ questions
  for machine comprehension of text,'' in \emph{EMNLP}, 2016.

\bibitem{Rajpurkar2018KnowWY}
P.~Rajpurkar, R.~Jia, and P.~Liang, ``Know what you don’t know: Unanswerable
  questions for squad,'' in \emph{ACL}, 2018.

\bibitem{Trischler2017NewsQAAM}
A.~Trischler, T.~Wang, X.~Yuan, J.~Harris, A.~Sordoni, P.~Bachman, and
  K.~Suleman, ``Newsqa: A machine comprehension dataset,'' in
  \emph{Rep4NLP@ACL}, 2017.

\bibitem{Dunn2017SearchQAAN}
M.~Dunn, L.~Sagun, M.~Higgins, V.~U. G{\"u}ney, V.~Cirik, and K.~Cho,
  ``Searchqa: A new q\&a dataset augmented with context from a search engine,''
  \emph{ArXiv}, vol. abs/1704.05179, 2017.

\bibitem{Joshi2017TriviaQAAL}
M.~Joshi, E.~Choi, D.~S. Weld, and L.~Zettlemoyer, ``Triviaqa: A large scale
  distantly supervised challenge dataset for reading comprehension,'' in
  \emph{ACL}, 2017.

\bibitem{Kwiatkowski2019NaturalQA}
T.~Kwiatkowski, J.~Palomaki, O.~Redfield, M.~Collins, A.~P. Parikh, C.~Alberti,
  D.~Epstein, I.~Polosukhin, J.~Devlin, K.~Lee, K.~Toutanova, L.~Jones,
  M.~Kelcey, M.-W. Chang, A.~M. Dai, J.~Uszkoreit, Q.~V. Le, and S.~Petrov,
  ``Natural questions: A benchmark for question answering research,''
  \emph{Transactions of the Association for Computational Linguistics}, vol.~7,
  pp. 453--466, 2019.

\bibitem{nguyen2016ms}
T.~Nguyen, M.~Rosenberg, X.~Song, J.~Gao, S.~Tiwary, R.~Majumder, and L.~Deng,
  ``Ms marco: A human generated machine reading comprehension dataset,'' in
  \emph{CoCo@ NIPS}, 2016.

\bibitem{Kocisk2018TheNR}
T.~Kocisk{\'y}, J.~Schwarz, P.~Blunsom, C.~Dyer, K.~Hermann, G.~Melis, and
  E.~Grefenstette, ``The narrativeqa reading comprehension challenge,''
  \emph{Transactions of the Association for Computational Linguistics}, vol.~6,
  pp. 317--328, 2018.

\bibitem{He2018DuReaderAC}
W.~He, K.~Liu, J.~Liu, Y.~Lyu, S.~Zhao, X.~Xiao, Y.~Liu, Y.~Wang, H.~Wu,
  Q.~She, X.~Liu, T.~Wu, and H.~Wang, ``Dureader: a chinese machine reading
  comprehension dataset from real-world applications,'' in \emph{QA@ACL}, 2018.

\bibitem{Xiong2019TWEETQAAS}
W.~Xiong, J.~Wu, H.~Wang, V.~Kulkarni, M.~Yu, S.~Chang, X.~Guo, and W.~Y. Wang,
  ``Tweetqa: A social media focused question answering dataset,'' in
  \emph{ACL}, 2019.

\bibitem{Rohrbach2015ADF}
A.~Rohrbach, M.~Rohrbach, N.~Tandon, and B.~Schiele, ``A dataset for movie
  description,'' \emph{2015 IEEE Conference on Computer Vision and Pattern
  Recognition (CVPR)}, pp. 3202--3212, 2015.

\bibitem{Baudis2015ModelingOT}
P.~Baudis and J.~Sediv{\'y}, ``Modeling of the question answering task in the
  yodaqa system,'' in \emph{CLEF}, 2015.

\bibitem{Dhingra2017QuasarDF}
B.~Dhingra, K.~Mazaitis, and W.~W. Cohen, ``Quasar: Datasets for question
  answering by search and reading,'' \emph{ArXiv}, vol. abs/1707.03904, 2017.

\bibitem{Guu2020REALMRL}
K.~Guu, K.~Lee, Z.~Tung, P.~Pasupat, and M.-W. Chang, ``Realm:
  Retrieval-augmented language model pre-training,'' \emph{ArXiv}, vol.
  abs/2002.08909, 2020.

\bibitem{karpukhin-etal-2020-dense}
\BIBentryALTinterwordspacing
V.~Karpukhin, B.~Oguz, S.~Min, P.~Lewis, L.~Wu, S.~Edunov, D.~Chen, and W.-t.
  Yih, ``Dense passage retrieval for open-domain question answering,'' in
  \emph{Proceedings of the 2020 Conference on Empirical Methods in Natural
  Language Processing (EMNLP)}.\hskip 1em plus 0.5em minus 0.4em\relax Online:
  Association for Computational Linguistics, Nov. 2020, pp. 6769--6781.
  [Online]. Available: \url{https://aclanthology.org/2020.emnlp-main.550}
\BIBentrySTDinterwordspacing

\bibitem{Lewis2020RetrievalAugmentedGF}
P.~Lewis, E.~Perez, A.~Piktus, F.~Petroni, V.~Karpukhin, N.~Goyal, H.~Kuttler,
  M.~Lewis, W.~tau Yih, T.~Rockt{\"a}schel, S.~Riedel, and D.~Kiela,
  ``Retrieval-augmented generation for knowledge-intensive nlp tasks,''
  \emph{ArXiv}, vol. abs/2005.11401, 2020.

\bibitem{roberts-etal-2020-much}
\BIBentryALTinterwordspacing
A.~Roberts, C.~Raffel, and N.~Shazeer, ``How much knowledge can you pack into
  the parameters of a language model?'' in \emph{Proceedings of the 2020
  Conference on Empirical Methods in Natural Language Processing
  (EMNLP)}.\hskip 1em plus 0.5em minus 0.4em\relax Online: Association for
  Computational Linguistics, Nov. 2020, pp. 5418--5426. [Online]. Available:
  \url{https://aclanthology.org/2020.emnlp-main.437}
\BIBentrySTDinterwordspacing

\bibitem{Berant2013SemanticPO}
J.~Berant, A.~K. Chou, R.~Frostig, and P.~Liang, ``Semantic parsing on freebase
  from question-answer pairs,'' in \emph{EMNLP}, 2013.

\bibitem{Miller2016KeyValueMN}
A.~H. Miller, A.~Fisch, J.~Dodge, A.-H. Karimi, A.~Bordes, and J.~Weston,
  ``Key-value memory networks for directly reading documents,'' in
  \emph{EMNLP}, 2016.

\bibitem{seo-etal-2019-real}
\BIBentryALTinterwordspacing
M.~Seo, J.~Lee, T.~Kwiatkowski, A.~Parikh, A.~Farhadi, and H.~Hajishirzi,
  ``Real-time open-domain question answering with dense-sparse phrase index,''
  in \emph{Proceedings of the 57th Annual Meeting of the Association for
  Computational Linguistics}.\hskip 1em plus 0.5em minus 0.4em\relax Florence,
  Italy: Association for Computational Linguistics, Jul. 2019, pp. 4430--4441.
  [Online]. Available: \url{https://aclanthology.org/P19-1436}
\BIBentrySTDinterwordspacing

\bibitem{wang-etal-2019-multi}
\BIBentryALTinterwordspacing
Z.~Wang, P.~Ng, X.~Ma, R.~Nallapati, and B.~Xiang, ``Multi-passage {BERT}: A
  globally normalized {BERT} model for open-domain question answering,'' in
  \emph{Proceedings of the 2019 Conference on Empirical Methods in Natural
  Language Processing and the 9th International Joint Conference on Natural
  Language Processing (EMNLP-IJCNLP)}.\hskip 1em plus 0.5em minus 0.4em\relax
  Hong Kong, China: Association for Computational Linguistics, Nov. 2019, pp.
  5878--5882. [Online]. Available: \url{https://aclanthology.org/D19-1599}
\BIBentrySTDinterwordspacing

\bibitem{izacard-grave-2021-leveraging}
\BIBentryALTinterwordspacing
G.~Izacard and E.~Grave, ``Leveraging passage retrieval with generative models
  for open domain question answering,'' in \emph{Proceedings of the 16th
  Conference of the European Chapter of the Association for Computational
  Linguistics: Main Volume}.\hskip 1em plus 0.5em minus 0.4em\relax Online:
  Association for Computational Linguistics, Apr. 2021, pp. 874--880. [Online].
  Available: \url{https://aclanthology.org/2021.eacl-main.74}
\BIBentrySTDinterwordspacing

\bibitem{Krishna2017DenseCaptioningEI}
R.~Krishna, K.~Hata, F.~Ren, L.~Fei-Fei, and J.~C. Niebles, ``Dense-captioning
  events in videos,'' \emph{2017 IEEE International Conference on Computer
  Vision (ICCV)}, pp. 706--715, 2017.

\bibitem{Heilbron2015ActivityNetAL}
F.~C. Heilbron, V.~Escorcia, B.~Ghanem, and J.~C. Niebles, ``Activitynet: A
  large-scale video benchmark for human activity understanding,'' \emph{2015
  IEEE Conference on Computer Vision and Pattern Recognition (CVPR)}, pp.
  961--970, 2015.

\bibitem{Rohrbach2016MovieD}
A.~Rohrbach, A.~Torabi, M.~Rohrbach, N.~Tandon, C.~Pal, H.~Larochelle, A.~C.
  Courville, and B.~Schiele, ``Movie description,'' \emph{International Journal
  of Computer Vision}, vol. 123, pp. 94--120, 2016.

\bibitem{Speer2017ConceptNet5A}
R.~Speer, J.~Chin, and C.~Havasi, ``Conceptnet 5.5: An open multilingual graph
  of general knowledge,'' in \emph{AAAI}, 2017.

\bibitem{Burton2009TheI2}
K.~R. Burton, A.~Java, and I.~Soboroff, ``The icwsm 2009 spinn3r dataset,''
  2009.

\bibitem{Lin2004ROUGEAP}
C.-Y. Lin, ``Rouge: A package for automatic evaluation of summaries,'' in
  \emph{ACL 2004}, 2004.

\bibitem{Papineni2002BleuAM}
K.~Papineni, S.~Roukos, T.~Ward, and W.-J. Zhu, ``Bleu: a method for automatic
  evaluation of machine translation,'' in \emph{ACL}, 2002.

\bibitem{banerjee-lavie-2005-meteor}
\BIBentryALTinterwordspacing
S.~Banerjee and A.~Lavie, ``{METEOR}: An automatic metric for {MT} evaluation
  with improved correlation with human judgments,'' in \emph{Proceedings of the
  {ACL} Workshop on Intrinsic and Extrinsic Evaluation Measures for Machine
  Translation and/or Summarization}.\hskip 1em plus 0.5em minus 0.4em\relax Ann
  Arbor, Michigan: Association for Computational Linguistics, Jun. 2005, pp.
  65--72. [Online]. Available: \url{https://aclanthology.org/W05-0909}
\BIBentrySTDinterwordspacing

\bibitem{Zhang2020BERTScoreET}
T.~Zhang, V.~Kishore, F.~Wu, K.~Q. Weinberger, and Y.~Artzi, ``Bertscore:
  Evaluating text generation with bert,'' \emph{ArXiv}, vol. abs/1904.09675,
  2020.

\bibitem{Sellam2020BLEURTLR}
T.~Sellam, D.~Das, and A.~P. Parikh, ``Bleurt: Learning robust metrics for text
  generation,'' in \emph{ACL}, 2020.

\end{thebibliography}
\bibliographystyle{IEEEtran}


\begin{IEEEbiography}[{\includegraphics[width=1in,height=1.25in,clip,keepaspectratio]{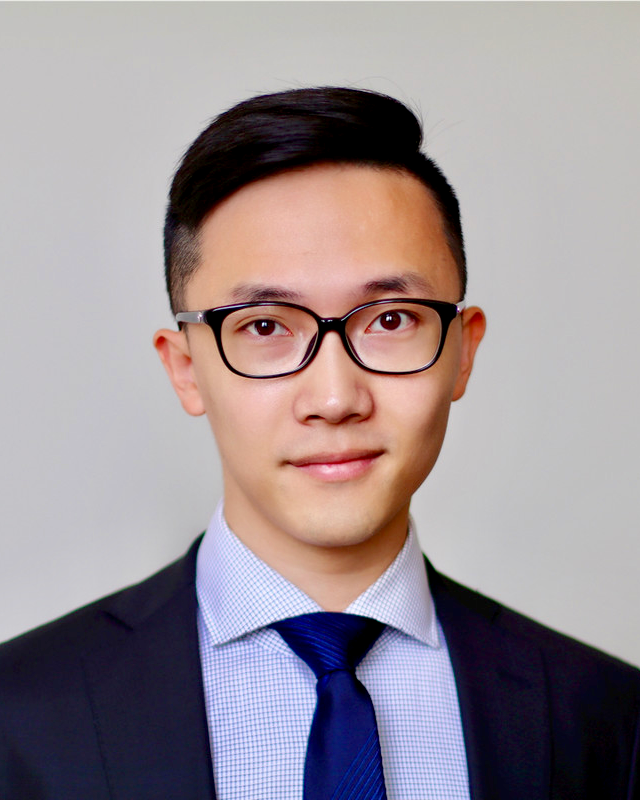}}]{Yang Bai}
received his M.S. degree from the ECE department at the University of Florida in 2018 and his Bachelor’s degree from the Microelectronics Department at the Sichuan University, China, in 2016. He is currently pursuing a PhD degree in computer science at the University of Florida under the supervision of Dr. Daisy Zhe Wang. His research interests are machine learning, natural language processing, and knowledge mining.
\end{IEEEbiography}

\begin{IEEEbiography}[{\includegraphics[width=1in,height=1.25in,clip,keepaspectratio]{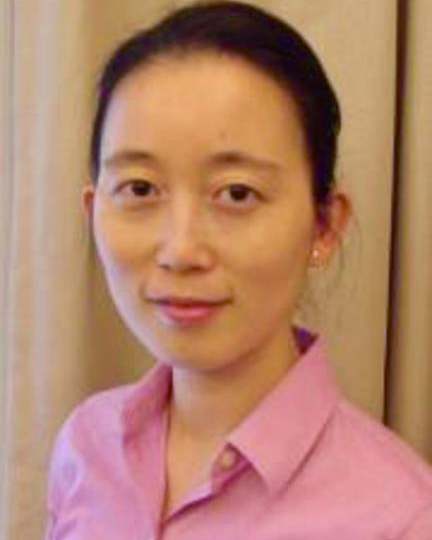}}]{Daisy Zhe Wang}
is an Associate Professor in the CISE department at the University of Florida. She is the Director of the Data Science Research Lab at UF. She obtained her Ph.D. degree from the EECS Department at the University of California, Berkeley in 2011 and her Bachelor´s degree from the ECE Department at the University of Toronto in 2005 . At Berkeley, she was a member of the Database Group and the AMP/RAD Lab. She is particularly interested in bridging scalable data management and processing systems with probabilistic models and statistical methods. She currently pursues research topics such as probabilistic databases, probabilistic knowledge bases, large-scale inference engines, query-driven interactive machine learning, and crowd assisted machine learning . She received Google Faculty Award in 2014. Her research is currently funded by NSF, DARPA, NIST, NIH, Google, Amazon, Pivotal, Greenplum/EMC, Sandia National Labs and Harris Corporation.
\end{IEEEbiography}




\end{document}